\documentclass[10pt]{article} %
\usepackage[preprint]{tmlr}

\usepackage{hyperref}
\usepackage{url}
\usepackage[utf8]{inputenc} %
\usepackage[T1]{fontenc}    %
\usepackage{hyperref}       %
\usepackage{url}            %
\usepackage{booktabs}       %
\usepackage{amsfonts}       %
\usepackage{nicefrac}       %
\usepackage{microtype}      %
\usepackage{xcolor}         %
\usepackage{makecell}
\usepackage{tcolorbox}

\usepackage{listings}
\usepackage{float}
\usepackage{fancyhdr}
\usepackage{graphicx}
\usepackage{amsmath}
\usepackage{amsthm}
\usepackage{amssymb}
\usepackage{listing}
\usepackage{enumitem}
\usepackage{apptools}
\usepackage{chngcntr}
\usepackage{adjustbox}
\usepackage{multirow}
\setcitestyle{citesep={;}}

\usepackage{mathtools}
\usepackage{wrapfig}
\usepackage{color}
\usepackage[ruled, vlined, linesnumbered]{algorithm2e}
\DeclareMathOperator*{\argmin}{argmin}
\DeclareMathOperator*{\argmax}{argmax}

\usepackage[capitalise,nameinlink,sort]{cleveref}
\Crefname{equation}{Eq.}{Eqs.}
\Crefname{assumption}{Assumption}{Assumptions}
\Crefname{condition}{Condition}{Conditions}
\Crefname{problem}{Linear Program}{Linear Programs}

\usepackage{quoting}

\usepackage[parfill]{parskip}

\usepackage{tikz}

\tikzstyle{round}=[thick,draw=black,circle]

\usepackage{thm-restate}
\DeclareFontFamily{U}{mathx}{\hyphenchar\font45}
\DeclareFontShape{U}{mathx}{m}{n}{<-> mathx10}{}
\DeclareSymbolFont{mathx}{U}{mathx}{m}{n}
\DeclareMathAccent{\widebar}{0}{mathx}{"73}

\theoremstyle{plain}
\newtheorem{theorem}{\textbf{Theorem}}

\newtheorem{lemma}{\textbf{Lemma}}

\newtheorem{proposition}[lemma]{Proposition}

\newtheorem*{lemma-a}{\textbf{Lemma}}
\newtheorem*{theorem-a}{\textbf{Theorem}}

\newtheorem{example}{Example}
\newtheorem{definition}{Definition}
\newtheorem{assumption}{Assumption}

\newtheorem{remark}{Remark}
\hypersetup{colorlinks=true,citecolor=blue,linkcolor=red}

\newcommand{\Acal}{\mathcal A}
\newcommand{\Bcal}{\mathcal B}

\newcommand{\Dcal}{\mathcal D}

\newcommand{\Fcal}{\mathcal F}

\newcommand{\Hcal}{\mathcal H}

\newcommand{\Lcal}{\mathcal L}
\newcommand{\Mcal}{\mathcal M}

\newcommand{\Ocal}{\mathcal O}
\newcommand{\Pcal}{\mathcal P}
\newcommand{\Qcal}{\mathcal Q}

\newcommand{\Scal}{\mathcal S}
\newcommand{\Tcal}{\mathcal T}

\newcommand{\Xcal}{\mathcal X}
\newcommand{\Ycal}{\mathcal Y}

\newcommand{\oec}{\textup{\texttt{OEC}}}
\newcommand{\lli}{\textup{\texttt{LL}}}
\newcommand{\HH}{\textup{\texttt{H}$^2$}}

\newcommand{\eoec}{\textup{\texttt{EOEC}}}

\newcommand{\TV}{\textup{\texttt{TV}}}
\newcommand{\KL}{\textup{\texttt{KL}}}
\newcommand{\RL}{\texttt{rl}}
\newcommand{\SL}{\texttt{sl}}

\renewcommand{\bar}{\widebar}
\renewcommand{\hat}{\widehat}
\renewcommand{\tilde}{\widetilde}

\let\sup\max
\let\inf\min

\title{A Fast Convergence Theory for Offline Decision Making}

\author{\name Chenjie Mao \email maochenjie@pjlab.org.cn\\
      \addr Shanghai Artificial Intelligence Laboratory\\
      \AND
\name Qiaosheng Zhang \email zhangqiaosheng@pjlab.org.cn\\
\addr Shanghai Artificial Intelligence Laboratory
}

\begin{document}

\maketitle

\begin{abstract}

    This paper proposes the first generic fast convergence result in general function approximation for offline decision making problems, which include offline reinforcement learning (RL) and off-policy evaluation (OPE) as special cases.
    To unify different settings, we introduce a framework called \emph{Decision Making with Offline Feedback} (DMOF), which captures a wide range of offline decision making problems.
    Within this framework, we propose a simple yet powerful algorithm called Empirical Decision with Divergence (EDD), whose upper bound can be termed as a coefficient named Empirical Offline Estimation Coefficient (EOEC).
    We show that EOEC is instance-dependent and actually measures the correlation of the problem.
    When assuming partial coverage in the dataset, EOEC will reduce in a rate of $1/N$ where $N$ is the size of the dataset, endowing EDD with a fast convergence guarantee.
Finally, we complement the above results with a lower bound in the DMOF framework, which further demonstrates the soundness of our theory.
\end{abstract}

\section{Introduction}

\emph{Offline decision making} studies learning decisions from a logged dataset that, in most cases, is only \emph{partially correlated} with the learning objective.
The possible partial correlation between the dataset and the learning objective distinguishes offline decision making from \emph{supervised learning}~\citep{Vapnik1998,books/daglib/0033642}, making it 
a promising subject capturing a wide range of real-world problems.
Examples of offline decision making include \emph{offline reinforcement learning}~(\citet{Antos2006LearningNP,Lange2012BatchRL,Levine2020OfflineRL}; offline RL)
and  \emph{off-policy evaluation}~(\citet{Li_2011,thomas2016dataefficient,farajtabar2018robust,uehara2022review}; OPE),  both of which are trendy areas in recent years.
Specifically, in the former,
the dataset is a logged interaction between the environment and the so-called behaviour policy while the 
learning objective is to learn a good policy that differs from the behaviour policy.
In the latter, the dataset is usually generated from a policy differing from the one to be estimated.
A crucial consideration in the offline decision making problem is \emph{learnability}, which refers to the ability to learn a desired policy from the logged dataset.

On one hand, although research in offline decision making has advanced rapidly, most theoretical results about learnability have been tied to specific problem structures, such as Markovian transition, tabular state-action space~(e.g., tabular Markov decision processes~(MDPs~\citep{Puterman1994MarkovDP}); \citet{Rashidinejad2021BridgingOR,Xie2021PolicyFB,Li2022SettlingTS}), and low-rank features~(e.g., linear MDPs; \citet{Jin2020IsPP,Huang2023ReinforcementLI,xiong2023nearly}), with assumptions like independently sampled data from a data distribution covering some policies~\citep{Chen2019InformationTheoreticCI,Liu2020ProvablyGB,Xie2020BatchVA,Xie2021BellmanconsistentPF}.
These results often differ in their formulations and rely on different assumptions for derivation. 
This diversity in problem structures and underlying assumptions not only obstructs a unified understanding of the offline decision making problem but also poses challenges in transferring theoretical results from one problem to another. %
Given these issues, we would like to ask:
\begin{center}
    \emph{Can we study offline decision making in a unified framework and extract a measure summarizing the learnability of offline decision making problems?}
\end{center}

On the other hand, the performance of algorithms in offline decision making in real-world scenarios can often exceed their theoretical results~\citep{kumar2020conservativeqlearningofflinereinforcement,levine2020offlinereinforcementlearningtutorial}, suggesting the potential for more refined results in this area. This is possibly due to the minimax nature of theoretical analysis, which takes undesired instances into consideration. Thus, a question that arises is: 
\begin{center}
    \emph{Can we derive an instance-dependent result with faster-convergence potential for offline decision making problems?}
\end{center}

This paper positively answers the two posed questions. We first introduce a unified framework for offline decision making, named \emph{Decision Making with Offline Feedback}~(DMOF).
We then present an algorithm that achieves instance-dependent regret as well as fast convergence rates in $1/N$ where $N$ is the sample size.
To complement our result, we further propose a lower bound for offline decision making problems. %
We detail our contributions as follows.
\begin{enumerate}
    \item We present a simple yet powerful algorithm named Empirical Decision with Divergence~(EDD; \Cref{eqn:ed2}), which enjoys the following advantages.
        \begin{itemize}
            \item (\Cref{theorem:up1}) It has an \emph{instance-dependent} upper bound that can adjust to different presentations of the real model. %
            \item (\Cref{theorem:fast-sq}) It attains a fast convergence rate~(specifically, a rate of 1/N, where N is the sample size) to the sub-optimality gap in 
                Markovian sequential problems~(\Cref{definition:s-p}) with partial coverage, where the latter includes Markov decision processes as a special case.
        \end{itemize}
    \item Next, we propose a measure---called \emph{Offline Estimation Coefficient}~($\oec$, \Cref{definition:oec})---that potentially encapsulates the hardness of an offline decision making problem. On top of this, we 
        provide an information-theoretic lower bound for the regret of any algorithm, expressed as a function of the $\oec$~(\Cref{theorem:lbexp}).
        We compare this lower bound with the upper bound achieved by EDD, demonstrating the soundness of EDD from lower bound's perspective.
\end{enumerate}

A summary of the main contributions of this paper is presented in \Cref{table:introduction}.

\begin{table*}[t]
  \centering
  \caption{A summary of the main contributions of this paper and results from similar model-based offline setting. The ``Assumptions'' column lists assumptions required by the result other than the DMOF framework~(\Cref{section:dmof}). ``Inst.'' means instance-dependent. ``Sequential'' means Markovian sequential problems~(\Cref{definition:s-p}). 
  The entries in the ``Value'' column are the bounds of the loss~(or return for \Cref{prop:2}). We strip some unnecessary terms away for simplicity.}\label{table:introduction}
\vspace{1em}
{\fontsize{6.5}{6.5}\selectfont
\begin{tabular}{cccccc}
\toprule
\multicolumn{2}{c}{Type} & Value/Gap & Location & Assumptions & Description\\
\midrule
\multirow{8}{*}{\makecell{\texttt{Upper Bounds}\\\texttt{of EDD}}} & \textup{Inst.} & $\min_{\lambda\ge 0}\eoec_{\lambda}(M^\star, \Mcal, \Dcal^\star)$ &  \Cref{theorem:up1} & \texttt{None} \\ \cmidrule{2-6} 
\cmidrule{2-5}
  &  \texttt{Minimax} & $3\Lcal(M, \bar{\pi})+C_{\bar{\pi}}\log\big(|\Mcal|/\delta\big)/N$ & \Cref{theorem:fast-sq} & \makecell{\texttt{Sequential},\\\Cref{assumption:s-iid,assumption:coverage}} & \multirow{6}{*}{\makecell{\texttt{directly}\\\texttt{follows from}\\\Cref{theorem:up1}}} \\\cmidrule{2-5}
&  \texttt{Minimax} & $J_{M_{\RL}^\star}^\star-C_{\pi^\star}\log\big(|\Mcal|/\delta\big)/N$ & \Cref{prop:2} & \makecell{\texttt{RL},\\\Cref{assumption:s-iid,assumption:coverage}} \\ \cmidrule{2-5}
&  \texttt{Minimax} & $\min_{\lambda\ge 0}[\oec_{\lambda}(\Mcal, \Mcal)+2\lambda\cdot\log(|\Mcal|/\delta)]$ & \Cref{theorem:up2} & \texttt{None} &\\
\midrule
\multicolumn{2}{c}{\texttt{Lower Bound}} & $\oec_{4B}(\Mcal, A)$ & \Cref{theorem:lbexp} & \texttt{None} \\
\midrule
\makecell{\texttt{Upper Bounds}\\\texttt{of \citet{Uehara2021PessimisticMO}}\\ \texttt{and \citet{bhardwaj2023adversarial}}}
&  \texttt{Minimax} & $J_{M_{\RL}^\star}^\star-\sqrt{C_{\pi^\star}\log\big(|\Mcal|/\delta\big)/N}$ & / & \makecell{\texttt{RL},\\\Cref{assumption:s-iid,assumption:coverage}} & \\
\bottomrule
\end{tabular}
}
\end{table*}

\subsection{Organization of This Paper}

We first introduce the concept of $f$-divergence in \Cref{section:preliminaries}, and then present the DMOF framework for offline decision making in \Cref{section:dmof}. 
After that, we propose the EDD algorithm in \Cref{section:ed2} and an instance-dependent upper bound for EDD. We further expand the upper bound in \Cref{section:fast} by showing a fast convergence result for Markovian sequential problems.
As an complement, \Cref{section:oec} introduces the hardness measure OEC and establishes a lower bound for offline decision making, and compares the previous upper bounds with it.
Finally, we conclude this paper with a discussion of the results and expectations for future work in \Cref{section:conclusion}.

\subsection{Related Works}

\paragraph{General Function Approximation in Offline Decision Making} General function approximation~\citep{bertsekas1996neuro,Vapnik1998,books/daglib/0033642} formulates the statistical learning problem with a generic function class, and studies the learnability of the problem in terms of the complexity of the function class. 
General function approximation is extensively studied in offline decision making, typically in the subjects like offline RL~\citep{Chen2019InformationTheoreticCI,Liu2020ProvablyGB,Xie2020BatchVA,Foster2021OfflineRL,Xie2021BellmanconsistentPF,Cheng2022AdversariallyTA,Zhan2022OfflineRL} and OPE~\citep{Uehara2019MinimaxWA,Jiang2020MinimaxVI,Huang2022BeyondTR}.
In these works, function classes are considered as another crucial aspect of learnability beyond data coverage.
Particular examples of these function classes include the value function class, the policy function class, the density ratio function class~\citep{Liu2018BreakingTC,Uehara2019MinimaxWA}, and the model class~\citep{Uehara2021PessimisticMO,bhardwaj2023adversarial}.
General function approximation is widely preferred since it is more adaptable to complex real-world scenarios~(both for function approximators and for the state-action space) and can be applied directly to specialized settings~(e.g., linear and tabular MDPs) with only minor modifications.
Recently, \citet{di2023pessimistic} proposes an algorithm named Pessimistic Nonlinear Least-Square Value Iteration for nonlinear function classes and get a minimax optimal instance-dependent regret when specialized to the linear setting. However, their results require uniform coverage over the dataset, which is fairly strong in the offline setting.

\paragraph{Instance-Dependent Bounds in Offline Decision Making}
Despite the rapid progress in the development of minimax bounds, instance-dependent analyses in offline decision making are still under development, possibly due to the challenge posed by partial correlation.
The most studied instance hardness measures are the variance and gap~\citep{xu2021fine,simchowitz2019non,he2021logarithmic} in the optimal value function, where the latter is defined as the difference of the values between the optimal action and sub-optimal ones across the state space.
Nevertheless, the gap-based instance-dependent bound seems less intuitive since it prefers problems with a large gap, though selecting sub-optimal actions with a small gap should also yield satisfactory results.
Specifically in the offline setting, there are only a few works presenting instance-dependent bounds, and all of them focus on special settings like linear MDPs~\citep{duan2020minimaxoptimal,yin2022nearoptimal,min2022varianceaware,nguyentang2023instancedependent} and tabular MDPs~\citep{pananjady2020instancedependent,yin2021instanceoptimal,wang2022gap}, with certain strong assumptions like uniform policy coverage~\citep{hu2021fast,yin2022offline,di2023pessimistic}.

\paragraph{Faster Convergence Rate in Sequential Problems} 
There are only a few works in offline decision making concerning faster convergence rate compared with $1/\sqrt{N}$, as far as we are aware. %
They either adapt the gap of optimal value function into the performance~\citep{hu2021fast,wang2022gap,nguyentang2023instancedependent}, or assume specific structures for the problem~(e.g., deterministic dynamics~\citep{uehara2021finite,yinnear}).
In contrast, our fast convergence applies to all Markovian sequential problems without the need for the gap, which yields a more satisfactory theory.

\paragraph{Works Related to DEC}
The framework and lower bound in this paper are influenced by the DEC paper for interactive~(online) decision making~\citep{foster2023statistical}, but it also introduces significant developments rather than being a simple adaptation.
For example, we use a fixed parameter $\lambda$, and we propose EDD that achieves an instance-dependent bound as well as a fast convergence rate for some particular problems such as supervised learning and Markovian sequential problems with partial coverage.
On the other hand, after the initial publication of \citet{foster2023statistical}, 
there have been subsequent works trying to enhance their result~\citep{foster2023modelfree,foster2023tight,chen2024unified} and expand it to other settings~\citep{foster2022complexity,foster2023complexity},
but none of these studies focuses on offline decision making. In conclusion, we hope that this part of our work can fill this gap and further inspire the development of DEC in the online setting.

\section{Preliminaries on $f$-Divergence}
\label{section:preliminaries}

We briefly introduce the concept of \emph{$f$-divergence} in this section, and refer the interested readers to Section $7$ in \citet{Polyanskiy2022Information} for a more detailed explanation.
We assume that we operate on a \emph{measurable space} $(E, \mathcal{E})$~(e.g., the set of \emph{real numbers} $\mathbb{R}$ and its \emph{Borel sets}), and we have access to a \emph{dominating measure} $\nu$~(e.g., the \emph{Lebesgue measure}).

For two probability measures $\mathbb{P}$ and $\mathbb{Q}$, 
we denote the \emph{Radon-Nikodym derivative} of $\mathbb{P}$ w.r.t. $\mathbb{Q}$ as $\frac{d\mathbb{P}}{d\mathbb{Q}}$.
We also use lowercase letters to denote the Radon-Nikodym derivative w.r.t. $\nu$~(e.g., $p\coloneqq d\mathbb{P}/d\nu$ and $q \coloneqq d\mathbb{Q}/d\nu$).

\begin{definition}[$f$-divergence]
    Given a function $f: (0, \infty)\to\mathbb{R}$ and two probability measures $\mathbb{P}$ and $\mathbb{Q}$, let $A\coloneqq\{x\in E|q(x)>0\}$ and $f^\prime(\infty)\coloneqq\lim_{x\to 0}xf(1/x)$,
    with the agreement that the $f^\prime(\infty)\mathbb{P}(E\setminus A)$ is $0$ if $\mathbb{P}(E\setminus A)=0$, 
    we define
    \begin{align*}
        D_f(\mathbb{P}\Vert \mathbb{Q})\coloneqq \int_A f\Big(\frac{p(x)}{q(x)}\Big)q(x)d\nu(x)+f^\prime(\infty)\mathbb{P}(E\setminus A)
    \end{align*}
    as the $f$-divergence between $\mathbb{P}$ and $\mathbb{Q}$ if $f$ is convex and $f(1)=0$.
\end{definition}

In short, the $f$-divergence measures the difference between two probability measures given a convex function $f$.
Examples of several frequently used $f$-divergences are listed below:
\begin{itemize}
    \item the \emph{KL divergence}, where $f(x)=x\log(x)$: 
            $\KL(\mathbb{P}\Vert \mathbb{Q})\coloneqq \mathbb{E}_{\nu}\big[p\log(p/q)\big]$;
    \item the \emph{squared Hellinger distance}, where $f(x)=(1-\sqrt{x})^2$: 
            $\texttt{H}^2(\mathbb{P}, \mathbb{Q})\coloneqq \mathbb{E}_{\nu}\big[(\sqrt{p}-\sqrt{q})^2\big]$;
    \item the \emph{total variation}, where $f(x)=0.5 | x-1|$: 
            $\TV(\mathbb{P}, \mathbb{Q})\coloneqq \mathbb{E}_{\nu}|p-q|/2$.
\end{itemize}

\section{Framework: Decision Making with Offline Feedback (DMOF)}
\label{section:dmof}

Inspired by the \emph{decision making with structured observation}~(DMSO) framework for online decision making problems~\citep{foster2023statistical},
we introduce a general framework for offline decision making, which is referred to as \emph{Decision Making with Offline Feedback}~(DMOF).
The DMOF framework includes an observation space $\Ocal$, a model space $\Mcal$, a decision space $\Pi$, and a loss function $\Lcal: \Mcal\times\Pi\to[0, B]$ for a constant $B > 0$.

The learner interacts with the DMOF according to the following protocol.
\begin{enumerate}
    \item \textbf{Problem Setup:} there is an unknown \emph{real model} $M^\star\in\Mcal$ specified ahead, under which we observe a dataset $\Dcal^\star\in\Ocal$ such that $\Dcal^\star\sim \mathbb{P}_{M^\star}$~(we will slightly abuse the notation and denote $\Dcal\sim M$ when considering the dataset generation).
    \item \textbf{Learning:} given the knowledge of $(\Mcal, \Pi, \Ocal, \Lcal)$ as well as the observed dataset $\Dcal^\star$, we wish to find 
        a decision $\hat{\pi}\in\Pi$ that minimizes the loss $\Lcal(M^\star, \hat{\pi})$. 
        We sometimes also learn a stochastic decision characterized by a distribution $p \in \Delta(\Pi)$, 
        of which the loss is defined as $\Lcal(M^\star, p)\coloneqq\mathbb{E}_{\pi\sim p}[\Lcal(M^\star,\pi)]$.
\end{enumerate}

This framework is carefully crafted for offline decision making problems, 
capturing broad settings such as offline RL, off-policy evaluation, and offline POMDPs.
\Cref{example:int_rl} provides a detailed explanation of how the DMOF framework is explicitly applied to the offline RL problem.

\begin{example}[DMOF in Offline RL]
    \label{example:int_rl}
    Let us consider the general framework of offline RL~(see \Cref{section:rl} for a more detailed introduction) with a realizable model class $\Mcal_{\RL}$ that contains $M_{\RL}^\star$.
    In such a setting, DMOF follows that: 
    $\Mcal=\{M=(M_{\RL}, d^\Dcal, N)| M_{\RL}\in\Mcal_{\RL}\}$ where $\Mcal_{\RL}$ is the set of MDPs, $d^\Dcal$ is the data distribution on the state-action space, and $N$ is the sample size; %
    $\Pi$ is the set of policies~($\Pi_{\RL}$); $\Ocal$ is the set of all possible dataset with size $N$;
    and $\Lcal((M_{\RL}, d^\Dcal, N), \pi)$ is the gap between the optimal expected return of $M_{\RL}\in\Mcal_{\RL}$ and 
    the expected return of policy $\pi$.
\end{example}

Algorithms~(denoted as $\mathfrak{A}$) are mappings from 
$(\Mcal, \Pi, \Ocal, \Lcal)$ and the observed dataset $\Dcal^\star$ to a
(distribution of) decision $\pi\in\Pi$. Since the elements $(\Mcal, \Pi, \Ocal, \Lcal)$ are always fixed, we will only specify $\Dcal^\star$ in most cases.
We now outline some advantages of the DMOF framework in the following.

\paragraph{$\Mcal$ as an Information Representor}
The complexity of $\Mcal$ reflects the information available in advance, and additional side information can help to prune $\Mcal$.
For any assumptions made or any information provided, we can always infer
a model class $\Mcal$ representing all necessary details.
For instance, in offline RL, assuming having a value function class $\Qcal$ that contains the optimal value function 
is equivalent to assuming having a model class containing the real model, which is composed of all possible models 
whose optimal value function is contained by $\Qcal$~\citep{mao2024on}. %

\paragraph{On the Variation of the Loss Function $\Lcal$}
It is worth noting that we do not put any constraints on the loss function $\Lcal$. It can be any function taking a model and a decision as input, and can even be irrelevant to its inputs.
Thus, one can specify $\Lcal$ to adapt to various scenarios. 
\emph{Essentially, the weak correlation between the loss function $\Lcal$ and the observation $\Dcal^\star$ forms one of the biggest differences between DMOF~(i.e., offline decision making) and standard supervised learning problems.}

\paragraph{On the Data Distribution} %
While prior works on offline decision making typically make many assumptions about the data generation process, 
these assumptions are made mainly for the ease of analysis and 
may not be necessary for learning a good policy in practice.
DMOF circumvents possible over-specification of data assumptions by the model class $\Mcal$, 
which, as we will show in the following sections, is
ultimately enough for summarizing the problem and its learnability.

\section{Algorithm: Empirical Decision with Divergence~(EDD)}
\label{section:ed2}

This section presents an algorithm named \emph{Empirical Decision with Divergence (EDD)} 
with an accompanying upper bound~(\Cref{theorem:up1}), which is instance-dependent and can be adapted to different real models.
Notably, this result is attained without further assumptions beyond the DMOF framework. As an expansion, we will further explain the guarantees of EDD under sequential learning problems like MDPs in \Cref{section:fast}, which ensures a fast convergence rate.
We defer the proofs for this section to \Cref{proof:ed2} for a clean presentation.

Given a dataset $\Dcal^\star$ generated by the real model $M^\star$, together with a model class $\Mcal$ standing as the information we can access, EDD outputs the policy in the following form: 
\begin{tcolorbox}[colback=white,colframe=black]%
\begin{align}
    \hat{p}(\Dcal^\star)= \argmin_{p\in\Delta(\Pi)} \max_{M\in\Mcal}\Big[\Lcal(M, p)+\lambda\log(\mathbb{P}_{M}(\Dcal^\star))\Big]\label{eqn:ed2}, 
\end{align}
\end{tcolorbox}
where $\lambda\ge 0$ is a hyper-parameter, and $\log(\mathbb{P}_M(\Dcal^\star))$ is the log-likelihood of observing $\Dcal^\star$ under model $M$.
\textcolor{black}{We will simply take $\hat{p}$ as the output distribution of policies when the context is clear.}
In short, EDD selects the policy that is universally good~(reflected by $\Lcal(M, \pi)$ in \Cref{eqn:ed2}) among possible real models~(reflected by $\log(\mathbb{P}_{M}(\Dcal^\star))$), but in a more unified way compared to the other classic model-based methods in RL~(e.g., \citet{Uehara2021PessimisticMO,foster2023statistical}), which first estimates a possible real model(s) and then tries to select a good policy.

\subsection{Instance-dependent Guarantee of EDD}

This subsection elaborates on the upper bounds achieved by EDD. Before that, we first introduce a new concept called the \emph{Empirical Offline Estimation Coefficient}~($\eoec$).

\begin{definition}
    \label{definition:eoec}
    Given a DMOF $(\Mcal, \Pi, \Ocal, \Lcal)$ and a dataset $\Dcal$, the $\eoec$ w.r.t. a model $M\in\Mcal$ and a parameter $\lambda\in\mathbb{R}$ is defined as 
\begin{align}
    \nonumber\eoec_{\lambda}(M, \Mcal, \Dcal)\coloneqq &
    \min_{p\in\Delta(\Pi)}\max_{M^\prime\in\Mcal} \big[\Lcal(M^\prime, p)\\& +\lambda\cdot\log(\mathbb{P}_{M^\prime}(\Dcal)/\mathbb{P}_{M}(\Dcal))\big].\label{eqn:eoec}
\end{align}
\end{definition}

We then state our first upper bound for EDD with the help of $\eoec$. 

\begin{theorem}
    \label{theorem:up1}
    For any real $M^\star\in\Mcal$ and $\Dcal^\star\sim M^\star$, EDD~(\Cref{eqn:ed2}) with a proper choice of its hyper-parameter has its loss bounded as 
    \begin{align}
        \Lcal(M^\star, \hat{p}(\Dcal^\star))\le \min_{\lambda\ge 0}\eoec_{\lambda}(M^\star, \Mcal, \Dcal^\star). \label{eqn:ub-i}
    \end{align}
\end{theorem}

As a function of both the real model and dataset,
this upper bound is instance-dependent and adapts automatically to various choices of $M^\star$ and even to the different occurrences of the dataset we encountered.
Although until now it may be unclear how tight the upper bound is, we will show in the next section that it is actually intriguing in many classic settings.

\section{Fast Convergence Rates for EDD under I.I.D. and Correlation}
\label{section:fast}

This section derives the fast convergence rate for offline decision making problems that are based on Markovian sequential problems, a typical example of which is the Markov decision process~(MDP)---the general framework of offline RL and OPE. %
We defer the proof of this section to \Cref{proof:fast}.

The key distinguishing property of a Markovian sequential problem is that its loss can be decomposed into a sequential form.
We formally define Markovian sequential problem in the following.

\begin{definition}[Markovian Sequential Problems]
    \label{definition:s-p}
    We say that a DMOF $(\Mcal, \Pi, \Ocal, \Lcal)$ is over a sequential problem~$(\Pcal, \Mcal, \{\Scal_h\}_{h\in[H]}, \{\Acal\}_{h\in[H]}, \Pi, \ell, N)$ if the following holds.
    \begin{enumerate}
        \item Given $\pi\in\Pi$ and $M\in\Mcal$, there exists a stochastic process $\{(s_h, a_h)\}_{h\in[H]}$ from the space $\Scal\otimes\Acal\coloneqq{(\Scal_1\times\Acal_1)\times (\Scal_2\times\Acal_2)\times\dots(\Scal_H\times\Acal_H)}$ such that $s_1\sim \Pcal(\cdot|M, 1)$, and for all $h\in[H]$, $a_h\sim\pi(\cdot |h, \{s_i, a_i\}_{i\in[h-1]}, s_h)$, $s_{h+1}\sim\Pcal(\cdot |M, h, s_h, a_h)$. %
            We denote the margin of $(s_h, a_h)$ w.r.t. the process as $\Tcal(\cdot |M, \pi, h)$, and the probability of the whole process as $\Tcal(\cdot| M, \pi)$.
        \item The loss is defined as $\Lcal(M, \pi)\coloneqq\mathbb{E}_{\{(s_h, a_h)\}_{h\in[H]}\sim \Tcal(\cdot| M, \pi)}\ell(\{(s_h, a_h)\}_{h\in[H]}, \pi)$, where $\ell\in((\Scal\otimes\Acal)\times \Pi\to[0, B])$.
    \end{enumerate}
    In such a case, we refer to $H\ge 1$ as the \emph{horizon} of this Markovian sequential problem. We will also add the subscript $M$ when referring to concepts specific to $M$.
\end{definition}

\begin{remark}[A Comparison Between Markovian Sequential Problems and MDPs]
    The Markovian sequential problem is a generalization of MDPs and has the following comprehensiveness:
    \begin{itemize}
        \item its policy is non-Markovian and can be based on the whole history; 
        \item instead of summing scalar rewards across the trajectory in MDPs, its loss~(namely, reversed return) can be any function on the whole history~(where the reward is merged into the state space $\Scal$). It can even be a function of the policy, which is typically useful in settings like risk-sensitive offline RL~\citep{urp2021riskaverse,rigter2023risk}.
    \end{itemize}
\end{remark}

Based on the above definition, we now make the i.i.d. assumption for \emph{the whole trajectories} in the dataset.
Note that we only assume i.i.d. across trajectories, while samples from the same trajectory can still be dependent.

\begin{assumption}[I.I.D. Assumption on a Sequential Problem]
    \label{assumption:s-iid}
    Given a DMOF $(\Mcal, \Pi, \Ocal, \Lcal)$ over a Markovian sequential problem $(\Pcal, \Mcal, \{\Scal_h\}_{h\in[H]}, \{\Acal_h\}_{h\in[H]}, \Pi, \ell, N)$, 
    the i.i.d. assumption assumes that 
    for all $M\in\Mcal$, the dataset $\Dcal\sim M$ consists of $N$ sequences $\{(s_{h, n}, a_{h, n}, s_{h, n}^\prime)\}_{h\in[H], n\in[N]}$ that are drawn in an i.i.d. manner, and for all $n\in[N]$ and $s_{h, n}^\prime\sim \Pcal(\cdot|M, \pi, s_{h, n}, a_{h, n})$. We denote the distribution of a sequence as $\Phi(M)$ and the margin of $(s_{h, n}, a_{h, n})$ as $d^\Dcal_h(\cdot|M)$.
\end{assumption}

\begin{remark}
    There are similar assumptions made for MDPs~(e.g., \citet{Rashidinejad2021BridgingOR,Xie2021BellmanconsistentPF,Cheng2022AdversariallyTA,Zhan2022OfflineRL,Huang2023ReinforcementLI}) in offline decision making literatures, which is a specification of \Cref{assumption:s-iid}.
    Some of them~(e.g., \citet{Li2022SettlingTS,xiong2023nearly,di2023pessimistic}) further require that $s_{h+1, n}=s_{h, n}^\prime$ or each trajectory from the dataset is an interaction generated by a fixed policy, which is more stringent compared with ours.
\end{remark}

We then define the coverage assumption, which confirms that the dataset can explain the behaviour of a policy.

\begin{assumption}[Coverage Assumption]
    \label{assumption:coverage}
    Given a DMOF $(\Mcal, \Pi, \Ocal, \Lcal)$ over a Markovian sequential problem $(\Pcal, \Mcal, \{\Scal_h\}_{h\in[H]}, \{\Acal_h\}_{h\in[H]}, \Pi, \ell, N)$ and real model $M^\star\in\Mcal$ that meet \Cref{assumption:s-iid}, 
    The coverage assumption assumes that there is a policy $\bar{\pi}\in\Pi$ and a constant $C_{\bar{\pi}}\ge 0$ such that 
    \begin{align}
        \max_{h\in[H]}\Big\lVert\frac{\Tcal(\cdot |M^\star, \bar{\pi}, h)}{d^\Dcal_h(\cdot| M^\star)}\Big\rVert_{\infty}\le C_{\bar{\pi}}.
    \end{align}
\end{assumption}

\begin{remark}
    The coverage assumption is standard in offline decision making, except for the tips explained in the following remark, and can be found in a wide range of works~(e.g., \citet{Rashidinejad2021BridgingOR,Chen2022OfflineRL,Zhan2022OfflineRL,Li2022SettlingTS,Huang2023ReinforcementLI}).
\end{remark}

\begin{remark}
    Throughout the paper, we do not assume that the models from $\Mcal$ share the same ``underlying distribution,'' which is widely done in offline decision making. On the other hand, we only assume the coverage assumption about $M^\star$, which makes our result more flexible.
\end{remark}

As a consequence of the above assumptions, we have the following fast convergence theorem for Markovian sequential problems.

\begin{theorem}
    \label{theorem:fast-sq}
    Given a DMOF $(\Mcal, \Pi, \Ocal, \Lcal)$ over a Markovian sequential problem $(\Pcal$, $\Mcal$, $\{\Scal_h\}_{h\in[H]}$, $\{\Acal_h\}_{h\in[H]}$, $\Pi$, $\ell, N)$ and $M^\star\in\Mcal$ that meet \Cref{assumption:s-iid} and \Cref{assumption:coverage} w.r.t. the policy $\bar{\pi}$ and the constant $C_{\bar{\pi}}$, with probability at least $1-\delta$ over $\Dcal^\star\sim M^\star$, 
    we have that 
    \begin{align*}
        &\eoec_{\frac{400BC_{\bar{\pi}}\log(2H)}{N}}(M^\star, \Mcal, \Dcal^\star)\\&\le
        3\Lcal(M^\star, \bar{\pi})+\frac{1200BC_{\bar{\pi}}\log(2H)}{N}\cdot\log\big(3H\cdot|\Mcal|/\delta\big).
    \end{align*}
    As a consequence, by choosing $\lambda$ as $\frac{400BC_{\bar{\pi}}\log(2H)}{N}$ in EDD, 
    with probability at least $1-\delta$ over $\Dcal^\star\sim M^\star$, EDD has that 
    \begin{align}
        \nonumber&\Lcal(M^\star, \hat{p})\\\le&
        3\Lcal(M, \bar{\pi})+\frac{1200BC_{\bar{\pi}}\log(2H)}{N}\cdot\log\big(3H\cdot|\Mcal|/\delta\big)\label{eqn:fast-sq2}.
    \end{align}
    Furthermore, if we have optimal coverage~(i.e., $\bar{\pi}=\pi^\star$) and $\Lcal(M^\star, \pi^\star)=0$, with probability at least  $1-\delta$  over $\Dcal\sim M^\star$, EDD has that 
    \begin{align}
        &\Lcal(M^\star, \hat{p})\le
        \frac{1200BC_{\pi^\star}\log(2H)}{N}\cdot\log\big(3H\cdot|\Mcal|/\delta\big)\label{eqn:fast-rl2}.
    \end{align}
\end{theorem}

\begin{remark}
    An inevitable concern about the above theorem is the term $3\Lcal(M^\star, \bar{\pi})$ appeared in \Cref{eqn:fast-sq2}, which shows that EDD may perform worse than the covered policy in some cases.
    We argue that our result is still impressive as EDD will always approach the optimal policy if the dataset covers it.
    This phenomenon---requiring the dataset to cover the optimal policy to perform well---can also be found in many works like \citet{Rashidinejad2021BridgingOR,Chen2022OfflineRL,Li2022SettlingTS}.
\end{remark}

We further specify the above theorem in offline RL in the following proposition, which is a direct consequence of it.

\begin{proposition}
\label{prop:2}
Let us consider the setting introduced in \Cref{section:rl},\footnote{We overload the notation as done in \Cref{example:int_rl}.} by setting $\Lcal(M, \pi)=J_{M_{\RL}}^\star-J_{M_\RL}(\pi)$, under the same context of \Cref{theorem:fast-sq}, by choosing $\lambda$ as $\frac{400BC_{\bar{\pi}}\log(2H)}{N}$, with probability at least $1-\delta$ over $\Dcal^\star\sim M^\star$, EDD has that 
    \begin{align}
        \nonumber J_{M^\star_{\RL}}^\star-J_{M^\star_{\RL}}(\hat{p})\le&
        3(J_{M^\star_{\RL}}^\star-J_{M^\star_{\RL}}(\bar{\pi}))\\&+
        \frac{1200BC_{\bar{\pi}}\log(2H)}{N}\cdot\log\big(3H\cdot|\Mcal|/\delta\big)\label{eqn:fast-rl1}.
    \end{align}
    If we have optimal coverage~(i.e., $\bar{\pi}=\pi^\star$), by choosing $\lambda$ as $\frac{400BC_{\bar{\pi}}\log(2H)}{N}$, with probability at least  $1-\delta$  over $\Dcal\sim M^\star$, EDD has that 
    \begin{align}
        J_{M^\star_{\RL}}^\star-J_{M^\star_{\RL}}(\hat{p})\le
        \frac{1200BC_{\pi^\star}\log(2H)}{N}\cdot\log\big(3H\cdot|\Mcal|/\delta\big)\label{eqn:fast-rl2}.
    \end{align}
\end{proposition}

\textcolor{black}{
\begin{remark}[Deriving a Result in OPE]
    We have not specified a result for OPE, but its adaptation is straightforward. The only requirement is to demonstrate that OPE represents a Markovian sequential problem: given a policy $\pi$ that we aim to estimate, the potential values of its return form the decision space ($\Pi$); the episodic loss function $\ell$, defined on model $M$ and a scalar value $r$ as an estimated return, is expressed as follows:
    \begin{align*}
        \ell(M, r)=
        \mathbb{E}_{\{(s_h, a_h)\}_{h\in[H]}\sim \Tcal(\cdot| M, \pi)}R_M(\{(s_h, a_h)\}_{h\in[H]})-r, 
    \end{align*}
    where $R_M$ denotes the return of the trajectory. 
    A rescale is probably needed to adjust the range of $\ell$. To get a result in terms of the absolute distance between $r$ and the real return, one can simply take $\ell(M, r) = -\ell(M, r)$ and apply \Cref{theorem:fast-sq} again
\end{remark}}

\paragraph{Comparison with Results in Model-based Offline RL with General Function Approximation} Our works in this subsection share similarities with \citet{Uehara2021PessimisticMO,bhardwaj2023adversarial}, but enjoy a substantial improvement. On the one hand, both their results converge in $\Ocal(\sqrt{\log(|\Mcal|)/N})$, which is the square root of ours.\footnote{Note that for a convergence, we would always want $\sqrt{\log(|\Mcal|)/N}\le 1$.} On the other hand, our setting is based on Markovian sequential problems, which is more general.

\subsection{Transferring Our Result to Toy Cases}

\textcolor{black}{
Unlike other general function approximation methods, model-based algorithms encounter difficulties when applied to toy cases, such as the tabular setting. This is primarily due to the introduction of an improper function class~(a function class that contains functions other than probability distributions) in general discretization techniques, such as covering numbers and bracketing numbers, which complicates their application in model-based algorithms. Specifically, the only existing method we are aware of for model-based algorithms is bracketing, as described in \citet{Uehara2021PessimisticMO}. 
However, this method may be flawed due to the use of two different definitions of the squared Hellinger divergence in an improper function class. A detailed discussion is provided in \Cref{section:bracketing}.}

This challenge is not entirely surprising: given that minimax bounds in these settings scale as $1/\sqrt{N}$, 
it would be unrealistic to expect our upper bound to surpass this limitation without introducing a new factor. Nonetheless, we believe this challenge is intriguing and represents a promising direction for future research. Specifically, it would be interesting to explore new assumptions, alternative methods, or problem-specific measures that could overcome this fundamental obstacle, allowing our results to be applied more smoothly to these cases and yield robust conclusions.

\section{Complementing Our Results with Lower Bounds}
\label{section:oec}

This section introduces the Offline Estimation Coefficient~($\oec$) and presents a lower bound~(\Cref{theorem:lbexp}) as a function of the $\oec$.
We will then compare the lower bound with the upper bound for EDD.
We defer the proofs for this section to \Cref{proof:oec}.

We define the $\oec$ as a trade-off between the similarities of the representation and the difference in learning objectives for models from DMOF, captured in a max-min-max form.
Compared with $\eoec$, $\oec$ replaces the ``data-based'' log-likelihood ratio with the squared Hellinger distance, leading us to remove the term ``empirical'' accordingly.

\begin{definition}
    \label{definition:oec}
    Given a DMOF $(\Mcal, \Pi, \Ocal, \Lcal)$, \emph{its} $\oec$ w.r.t. a set of probability $A\subseteq\Delta(\Ocal)$ and a parameter $\lambda \ge 0$ is defined as 
    \begin{align}
        \nonumber&\oec_{\lambda}(\Mcal, A)\\\coloneqq&
        \sup\limits_{\rho\in A}\inf\limits_{p\in\Delta(\Pi)}
        \sup\limits_{M\in\Mcal}\Big[\Lcal(M, p)-\lambda\cdot\HH(\rho, M)\Big].
        \label{eqn:oec}
    \end{align}
    where $\HH(\rho, M)$ is the squared Hellinger distance between $\rho$ and $\mathbb{P}_M$.
\end{definition}

Below, we provide an interpretation of the formulation and the operational meaning of $\oec$. Technically, the first
$\max_{\rho\in A}$ part in $\oec$ sets a point in the model class $\Mcal$ where the real model possibly lies, suggesting that the real model should be similar to $\rho$. The $\min_{p\in\Delta(\Pi)}$ indicates 
that there is a learner pessimistically chooses a stochastic policy $p \in \Delta(\Pi)$, with the knowledge of $\rho$ but without knowing the real model $M \in \mathcal{M}$.
The last $\max_{M\in\Mcal}$ part
means that an adversary selects the worst-case real model $M$ that is similar to $\rho$ after knowing the stochastic policy $p$.

The $\oec$ summarizes the learnability property of a DMOF problem: as will be presented in \Cref{theorem:lbexp}, it implies a lower bound for learning. Moreover, the $\oec$ is also involved in the upper bound presented in \Cref{theorem:up2}.

\begin{remark}
As a further development, the squared Hellinger distance in \Cref{eqn:eoec} can be replaced by other $f$-divergences 
provided they satisfy a refined change of measure property~(introduced in \Cref{section:extension}).
Such $f$-divergences include the total variation distance, the KL divergence, and so forth. We will discuss them in more detail in \Cref{section:extension}.
\end{remark}

\subsection{Lower Bounds via $\oec$ without Further Assumptions}
\label{subsection:lb}

This subsection presents a lower bound on the loss of
any algorithm, which is specifically expressed as a function of the $\oec$ with $\lambda=4B$.
Here, we do not make any assumption on the problem---e.g., assumptions on structure of the problem~(like sequential or Markovian) or on the coverage of the dataset---beyond the DMOF framework.
Our result on the lower bound is encapsulated in the following theorem.

\begin{theorem}
\label{theorem:lbexp}
Given a DMOF $(\Mcal, \Pi, \Ocal, \Lcal)$, 
for any algorithm $\mathfrak{A}$, there exists an $M\in\Mcal$ such that the expected loss of the algorithm under model $M$ is lower bounded as: 
\begin{align}
    \label{eqn:lb}
    \mathbb{E}_{\Dcal\sim M}\Big[\Lcal(M, \mathfrak{A}(\Dcal))\Big]\ge\frac{1}{3}\oec_{4B}(\Mcal, A), 
\end{align}
where $A$ can be chosen as any subset of $\Delta(\Ocal)$.
\end{theorem}

\begin{remark}[About Data Size]
    In contrast to normal conditions, there is no explicitly the size of dataset showed in this lower bound. 
    Inherently, the information of the size of the dataset is measured by the Hellinger distance in $\oec$.
\end{remark}

An interesting observation is
that the set $A$ can be any set of distributions on $\Ocal$, which, while of course including $\Mcal$ itself, are not limited to models from DMOF.
For instance,  in the context of time-inhomogenious Markov decision processes, offline RL often makes independent assumptions on samples~\citep{Chen2019InformationTheoreticCI,Liu2020ProvablyGB,Rashidinejad2021BridgingOR,Xie2021BellmanconsistentPF}, whereas $\Delta(\Ocal)$ can contain distributions that are correlated.
This somewhat contrasts with the typical approach to proving lower bounds: to find a lower bound from a problem set $\Mcal$, 
we typically construct a subset of $\Mcal$ and verify the hardness of the subset.

As discussed above, the purpose of the first $\max$ part is merely to serve as a guide for pinpointing the real model.
The lower bound becomes tightest when $A$ is taken as $\Delta(\Ocal)$, but computing the exact value of $\oec$ for $A=\Delta(\Ocal)$ may require significant effort. When computing the lower bound, one can choose an appropriate $A$ that balances accuracy with computational complexity.

As a complement, the following theorem relates $\eoec$ with $\oec$, showing that EDD is nearly optimal in the minimax perspective.

\begin{theorem}
    \label{theorem:up2}
    For any DMOF $(\Mcal, \Pi, \Ocal, \Lcal)$ and model $M\in\Mcal$, with probability at least $1-\delta$ over $\Dcal\sim M$, 
    \begin{align*}
        \eoec_{\lambda}(M, \Mcal, \Dcal)\le&\oec_{\lambda}(\Mcal, \{M\})+2\lambda\cdot\log(|\Mcal|/\delta).
    \end{align*}
    Consequently, for any DMOF $(\Mcal, \Pi, \Ocal, \Lcal)$ and real $M^\star\in\Mcal$, EDD~(\Cref{eqn:ed2}) with a proper choice of its hyper-parameters has its loss bounded, with probability at least $1-\delta$ over $\Dcal^\star\sim M^\star$, as 
    \begin{align}
        \nonumber\Lcal(M^\star, \hat{p}(\Dcal^\star))\le&\min_{\lambda\ge 0}\big[\oec_{\lambda}(\Mcal, \{M^\star\})+2\lambda\cdot\log(|\Mcal|/\delta)\big]\\\le&
        \min_{\lambda\ge 0}\big[\oec_{\lambda}(\Mcal, \Mcal)+2\lambda\cdot\log(|\Mcal|/\delta)\big], \label{eqn:ub-m}
    \end{align}
    where $|\Mcal|$ denotes the cardinality of $\Mcal$.
\end{theorem}

\begin{remark}
    The cardinality $|\Mcal|$ in the above theorem can be replaced by other complexity measures like the covering number and the Rademacher complexity with minor modifications.
\end{remark}

\subsection{Comparing the Upper Bound and the Lower Bound}
\label{subsection:comparing}

Apart from the $\oec$ term in the lower bound~(\Cref{eqn:lb}), the upper bound~(\Cref{eqn:ub-m}) also contains an additional $2\lambda\cdot\log(|\Mcal|)$. 
For a convergent result, one must set $\lambda$ to a sufficiently small value compared with the constant value in \Cref{eqn:lb}, which still makes sense as we will explain in the following.

Notably, until now, we have not made any assumptions such as ``independent and identically distributed'' (i.i.d.) or the coverage assumption, except for the framework depicted by DMOF.
Thus, we can expect that either the lower bound (\Cref{eqn:lb}) only occurs in extreme cases and may not accurately reflect the tightness of the upper bounds (\Cref{eqn:ub-i,eqn:ub-m}), or that the upper bounds can be significantly improved with certain appropriate assumptions.

Similar observations~(the additional $\lambda\cdot\log(|\Mcal|/\delta)$ and \textcolor{black}{the values of $\lambda$ in upper and lower bounds}) also appear in occur for the PAC bound in the online setting in the spectrum works of DEC~\citep{foster2023tight},  
which is regarded as an unsolved problem.
An interesting point about the regret bound in the online setting is that, while the addition term remains, we do not have the problem of setting $\lambda$ to be too small~\citep{foster2023statistical,foster2023tight}. This is because every distribution~(or interaction) has a loss appearing in the regret, which allows us to use a more moderate value about $\lambda$. We refer interested readers to the original papers for more details.

\section{Conclusion}
\label{section:conclusion}

This paper proposes the first generic fast convergence result for general function approximation.
We first introduce a general framework called the Decision Making with Offline Feedback~(DMOF) for offline decision making, 
under which we present an algorithm named EDD, which achieves an instance-dependent upper bound. 
We further enhance it with a fast convergence result under Markov sequential problems, which takes MDPs as a special case.
Following this, we introduce a complexity measure called the 
offline estimation coefficient~($\oec$), based on which we build an information-theoretic lower bound to complement our results.

Finally, we highlight some possible further directions from this paper.
Firstly, as our results in this paper are based on model-based general function approximation for general offline decision making problems, it would be interesting to concretize our results, extend them to special settings like POMDPs, and to find a way to reasonably apply it to toy cases like tabular settings.
Additionally, while part of this paper can be considered as an offline work influenced by \citet{foster2023statistical} in the online setting, the results presented here could also have online extensions
and may enhance the results for DEC.

\bibliography{/home/aithus/.latex/instance,/home/aithus/.latex/Eprocess,/home/aithus/.latex/ml,/home/aithus/.latex/ope,/home/aithus/.latex/ref,/home/aithus/.latex/online,/home/aithus/.latex/information_theory,/home/aithus/.latex/empirical}
\appendix

\section{A Short Introduction to Offline Reinforcement Learning}
\label{section:rl}

This section would shortly introduce offline reinforcement learning~(offline RL), i.e., learning with offline Markov decision processes~(MDP). To distinguish offline RL from DMOF, we would add the subscript $\RL$ when the notations collapse.

For the pre-learning interaction part, we consider the time-inhomogeneous MDP. Suppose that we have a MDP 
\begin{align*}
M_{\RL}\coloneqq(\{\Scal_h\}_{h\in[H]}, \{\Acal_h\}_{h\in[H]}, d_1, \{R_{h}\}_{h\in[H]}, \{P_{h}\}_{h\in[H]}, \{\Pi_{\RL, h}\}_{h\in[H]}), 
\end{align*}
where $\{\Scal_h\}_{h\in[H]}$ is the state spaces, $\{\Acal_h\}_{h\in[H]}$ is the action spaces, $d_1\in\Delta(\Scal_1)$ is the initial state distribution, $\{R_h\}_{h\in[H]}$ is the reward functions, $\{P_{h}\}_{h\in[H]}$ is the transition kernel, $\{\Pi_h\}_{h\in[H]}$ is the policy space, and $H$ is the horizon. 
For simplicity, we further denote $\Scal\coloneqq\{\Scal_h\}_{h\in[H]}$, $\Acal\coloneqq\{\Acal_h\}_{h\in[H]}$, $P\coloneqq\{P_h\}_{h\in[H]}$, and $\Pi_{\RL}\coloneqq\{\Pi_{\RL, h}\}_{h\in[H]}$.

In the model-based setting, there will be a model $\Mcal_{\RL}$ being a set of MDPs which only differ in $\{P_h\}_{h\in[H]}$ and $\{R_h\}_{h\in[H]}$~(mostly only $\{P_h\}_{h\in[H]}$).

Given a police $\pi\coloneqq\{\pi_{\RL, h}\}_{h\in[H]}$ where $\pi_{\RL, h}\in\Pi_{\RL, h}$ for $h\in[H]$ and a real model $M^\star_{\RL}$, we would interact with the MDP according to the following protocol. 

\begin{itemize}
    \item We first sample a initial state $s_1$ from $d_1$.
    \item for every $h\in[H]$, we would take the action $a_h\sim\pi_{\RL, h}(s_h)$, receive the reward $R_h^\star(s_h, a_h)$ and transit to the next state $s_{h+1}\sim P^\star_h(\cdot|s_h, a_h)$; at the step $h=H$, we will deterministically transfer to the ending state $s_{H+1}$. Both $R^\star$ and $P^\star$ come from $M_{\RL}^\star$.
\end{itemize}

Our goal is to maximize the cumulative reward, namely, $J_{M_{\RL}^\star}(\pi)\coloneqq\mathbb{E}\sum_{h\in[H]}R_h(s_h, a_h)$~(we will broadcast this definition to other $M_{\RL}\in\Mcal_{\RL}$).
We assume $J_{M_{\RL}}(\pi_{\RL})\in[0, B]$ for all $M_{\RL}\in\Mcal_{\RL}$ and $\pi_{\RL}\in\Pi_{\RL}$.
Given a distribution $p$ over $\Pi$, we define $J_{M_{\RL}}(p)\coloneqq\mathbb{E}_{\pi_{\RL}\sim p}J_{M_{\RL}}(pi_{\RL})$.
We further denote $\pi^\star_{\RL, M}\coloneqq\argmax_{\pi_{\RL}\in\Pi_{\RL}}J_{M_{\RL}}(\pi_{\RL})$ and $J^\star_{M_{\RL}}\coloneqq J_{M_{\RL}}(\pi^\star_{\RL, M})$.

For the offline learning part, we are given with the dataset $\Dcal_{\RL}^\star\coloneqq\{(s_{h, n}, a_{h, n}, r_{h, n}, s_{h, n}^\prime)\}_{h\in[H], n\in[N]}$ where $N$ is the number of samples, and aim to learn a policy $\pi_{\RL}$ to maximize $J_{M_{\RL}^\star}(\pi)$.
Typically we will have that $r_{h, n}=R_{h}^\star(s_{h, n}, a_{h, n})$ and $s^\prime_{h, n}\sim P_h^\star(\cdot|s_{h, n}, a_{h, n})$.%
\footnote{Note that some works also assume that $s_{h+1, n}=s_{h, n}^\prime$, which can be achieved by specifying the above framework.}
To gain a converging result, offline RL always makes the i.i.d. assumption, namely, there is a data distribution $d^\Dcal\coloneqq\{d^\Dcal_h\}_{h\in[H]}$ that for all $n\in[N]$, we have $(s_{h, n}, a_{h, n})\sim d^\Dcal_h$, and $(s_{h, n}, a_{h, n}, r_{h, n}, s_{h, n}^\prime)$ are independent of each other across $n\in[N]$ for all $h\in[H]$.

\section{Some Useful Results in Probability and Statistics}

\begin{lemma}
    \label{lemma:derivative}
    Given a measurable space $(\Omega, \mathcal{E})$, and two measures $\phi$ and $\nu$ such that $\phi$ is absolutely continuous w.r.t. $\nu$, 
    for any real valued measurable function $f$ on $\Omega$, 
    \begin{align*}
        \int f(x)\frac{d\phi(x)}{d\nu(x)} \nu(dx)=
        \int f(x) \phi(dx).
    \end{align*}
\end{lemma}

\begin{proof}
    We can prove the case that $f$ is \emph{simple}, and the extension to the other cases should be straightforward due to the definition of the Lebesgue integral.\footnote{We follow the definitions from \citet{Stroock2011EssentialsOI}.}

    Defining the set $\Bcal\coloneqq\{f^{-1}(y)| y\in\mathbb{R}\}$, we must have $\bigcup_{B\in\Bcal} = \Omega$.
    $\Bcal$ is countable and consists of disjoint measurable sets since $f$ is simple and measurable.
    For $B\in\Bcal$, we denote $f(B)$ as the value of $f$ for any element belonging to $B$.
    Due to the definition of the Lebesgue integral, 
    \begin{align*}
        \int f(x) \phi(dx)=&\sum\limits_{B\in\Bcal} f(B)\phi(B)\\=&
        \sum\limits_{B\in\Bcal} f(B)\int_{B}\frac{d\phi(x)}{d\nu(x)}\nu(dx)\\=&
        \sum\limits_{B\in\Bcal} \int_{B}f(x)\frac{d\phi(x)}{d\nu(x)}\nu(dx)\\=&
        \int f(x)\frac{d\phi(x)}{d\nu(x)} \nu(dx).
    \end{align*}

    The last equality is justified by the additivity of the Lebesgue integration~\citep{668958}.
    This completes the proof.
\end{proof}

\subsection{Change of Measure Properties}

One of the main properties of the $f$-divergence exploited in this paper~(as done in \citet{foster2023statistical}) is the \emph{change of measure} property.
Concretely, the change of measure property is concerned with changing the probability measure underlying an expectation, with the cost of a divergence between the probabilities, plus a possible variation on the function in the expectation.
The change of measure inequalities of $f$-divergences are essential in learning theory, especially in fields like PAC-Bayesian learning~\citep{ShaweTaylor1997APA,McAllester1998SomePT,McAllester1999PACBayesianMA,alquier2023userfriendly}.

We briefly state some results about change of measure properties in this subsection, and would like to refer interested readers to \citet{ohnishi2020novel} for a more comprehensive review of the change of measure inequalities.
Also note that this paper would focus on some refined change of measure properties, which would be explained in more detail in \Cref{section:extension}.

We first state the classic \emph{Donsker and Varadhan variational formula} in the following lemma.

\begin{lemma}[Donsker and Varadhan's Formula]
    \label{lemma:donsker}
Let $X$ be a real-valued random variable~(r.v.) over $\Xcal$, $h: \Xcal\to\mathbb{R}$ be a real-valued function, and $\rho, \pi$ be probability measures over $\Xcal$, we have that 
    \begin{align}
        \textup{KL}(\rho\rVert \pi) = \sup_{h}\Big[\mathbb{E}_{X\sim \rho}[h(X)] -\log\mathbb{E}_{X\sim \pi}[\exp(h(X))] \Big].
    \end{align}
\end{lemma}

The following shows the change of measure property of the total variation distance.

\begin{lemma}
    \label{lemma:tv-change}
    Given a measurable space $(\Omega, \mathcal{E})$, two probabilities $\mathbb{P}$ and $\mathbb{Q}$, and a positive number $\beta>0$, 
    for any real-valued measurable function $f$ on $\Omega$ with range $[0, \beta]$, 
    \begin{align*}
        \big|\mathbb{E}_{\mathbb{P}}[f] -\mathbb{E}_{\mathbb{Q}}[f]\big|
        \le2\beta\TV(\mathbb{P}, \mathbb{Q}).
    \end{align*}
\end{lemma}

\begin{proof}
    Take $\nu\coloneqq \mathbb{P}+\mathbb{Q}$ as the dominating measure, 
    we have that
    \begin{align*}
        \TV(\mathbb{P}, \mathbb{Q})=&\frac{\int\Big|\frac{d\mathbb{P}}{d\nu}-\frac{d\mathbb{Q}}{d\nu}\Big|d\nu}{2}\\=&
        \frac{\int\beta\Big|\frac{d\mathbb{P}(x)}{d\nu(x)}-\frac{d\mathbb{Q}(x)}{d\nu(x)}\Big|\nu(dx)}{2\beta}\\\ge&
        \frac{\int f(x)\Big|\frac{d\mathbb{P}(x)}{d\nu(x)}-\frac{d\mathbb{Q}(x)}{d\nu(x)}\Big|\nu(dx)}{2\beta}\\=&
        \frac{\int \Big|f(x)\frac{d\mathbb{P}(x)}{d\nu(x)}-f(x)\frac{d\mathbb{Q}(x)}{d\nu(x)}\Big|\nu(dx)}{2\beta}\\\ge&
        \tag{Jensen's inequality}
        \frac{\Big|\int \Big(f(x)\frac{d\mathbb{P}(x)}{d\nu(x)}-f(x)\frac{d\mathbb{Q}(x)}{d\nu(x)}\Big)\nu(dx)\Big|}{2\beta}\\=&
        \frac{\Big|\int \big(f(x)\frac{d\mathbb{P}(x)}{d\nu(x)}\big) \nu(dx)-\int\big(f(x)\frac{d\mathbb{Q}(x)}{d\nu(x)}\big)\nu(dx)\Big|}{2\beta}\\=&
        \frac{\big|\mathbb{E}_{\mathbb{P}}[f] -\mathbb{E}_{\mathbb{Q}}[f]\big|}{2\beta}\tag{\Cref{lemma:derivative}}
    \end{align*}
    The fifth inequality is due to Jensen's inequality and the fact that the absolute value function is convex.

    This completes the proof.
\end{proof}

Also, there is a similar result for the squared Hellinger distance.

\begin{lemma}[Trimmed Lemma A.11 in \citet{foster2023statistical}]
    \label{lemma:hh-change}
    Let $\mathbb{P}$ and $\mathbb{Q}$ be probability measures on $(\Xcal, \Fcal)$, for all $g:\Xcal\to[0, B]$, 
    \begin{align*}
        \mathbb{E}_{X\sim \mathbb{P}}[g(X)]\le 3\mathbb{E}_{X\sim\mathbb{Q}}[g(X)]+4B\cdot\textup{\texttt{H}}^2(\mathbb{P}, \mathbb{Q}).
    \end{align*}
\end{lemma}

\begin{proof}
    See \citet{foster2023statistical}.
\end{proof}

\section{Missing Proofs in \Cref{section:ed2}}
\label{proof:ed2}

\subsection{Some Helper Lemmas}

The following lemma presents a relation between the log-likelihood and the squared Hellinger distance.

\begin{lemma}[Union Bound of the Log-Likelihood]
    \label{lemma:ll}
    Let $\Pcal$ be a set of probability measures, and $\Dcal$ be a dataset where $\Dcal\sim p^\star$ for some $p^\star\in\Pcal$.
    Then, with probability at least $1-\delta$ over $\Dcal\sim p^\star$, for all $p\in\Pcal$, 
    \begin{align*}
        -\log(p^\star(\Dcal)/p(\Dcal))+
        \textup{\texttt{H}}^2(p^\star, p)\le 2\log(|\Pcal|/\delta).
    \end{align*}
\end{lemma}

\begin{proof}
    We first fix $p\in\Pcal$, and take 
    \begin{align*}
        L(\Dcal)\coloneqq
        -\frac{1}{2}\log(p^\star(\Dcal)/p(\Dcal))-
        \log{\mathbb{E}_{\Dcal^\prime\sim p^\star}\big[\sqrt{p(\Dcal^\prime)/p^\star(\Dcal^\prime)}\big]}.
    \end{align*}
    We have the moment generating function of $L(\Dcal)$ be $1$ at the value $1$, 
    \begin{align*}
        \mathbb{E}_{\Dcal\sim p^\star}\Big[\exp\big(L(\Dcal)\big)\Big]=&
        \mathbb{E}_{\Dcal\sim p^\star}\Big[\exp\big(
        -\frac{1}{2}\log(p^\star(\Dcal)/p(\Dcal))-
        \log{\mathbb{E}_{\Dcal^\prime\sim p^\star}\big[\sqrt{p(\Dcal^\prime)/p^\star(\Dcal^\prime)}\big]}
        \big)\Big]\\=&
        \frac{\mathbb{E}_{\Dcal\sim p^\star}\Big[\exp\big(-\frac{1}{2}\log(p^\star(\Dcal)/p(\Dcal))\big)\Big]}
        {\mathbb{E}_{\Dcal^\prime\sim p^\star}\big[\sqrt{p(\Dcal^\prime)/p^\star(\Dcal^\prime)}\big]}\\=&1.
    \end{align*}
    We can then apply the Chernoff bound, which yields that 
    \begin{align*}
        \mathbb{P}_{\Dcal\sim p^\star}\big(L(\Dcal)\ge\epsilon\big)\le&
        \frac{\mathbb{E}_{\Dcal\sim p^\star}\Big[\exp\big(L(\Dcal)\big)\Big]}{\exp(\epsilon)}=\exp(-\epsilon).
    \end{align*}
    Further, by the definition of the squared Hellinger distance, 
    \begin{align*}
        &\mathbb{P}_{\Dcal\sim p^\star}\big(
        -\log(p^\star(\Dcal)/p(\Dcal))
        +\textup{\texttt{H}}^2(p^\star, p)\ge\epsilon\big)\\=&
        \mathbb{P}_{\Dcal\sim p^\star}\big(
        -\log(p^\star(\Dcal)/p(\Dcal))+
        2-2\mathbb{E}_{\Dcal^\prime\sim p^\star}\big[\sqrt{p(\Dcal^\prime)/p^\star(\Dcal^\prime)}\big]\ge\epsilon\big)\\=&
        \mathbb{P}_{\Dcal\sim p^\star}\big(
        -\log(p^\star(\Dcal)/p(\Dcal))+
        2-2\exp\log\mathbb{E}_{\Dcal^\prime\sim p^\star}\big[\sqrt{p(\Dcal^\prime)/p^\star(\Dcal^\prime)}\big]\ge\epsilon\big)
    \end{align*}
    Due to the fact that $\exp(x)\ge x+1$ for $x\in\mathbb{R}$, we further have 
    \begin{align*}
        &\mathbb{P}_{\Dcal\sim p^\star}\big(
        -\log(p^\star(\Dcal)/p(\Dcal))+\textup{\texttt{H}}^2(p^\star, p)\ge\epsilon\big)\\\le&
        \mathbb{P}_{\Dcal\sim p^\star}\big(
        -\log(p^\star(\Dcal)/p(\Dcal))-
        2\log{\mathbb{E}_{\Dcal^\prime\sim p^\star}\big[\sqrt{p(\Dcal^\prime)/p^\star(\Dcal^\prime)}\big]}
        \ge\epsilon\big)\\=&
        \mathbb{P}_{\Dcal\sim p^\star}\big(L(\Dcal)\ge\epsilon/2\big)\\\le&\exp(-\epsilon/2).
    \end{align*}
Taking $\delta=\exp(-\epsilon/2)$, we have $\epsilon=2\log(1/\delta)$. 
We can conclude that with probability at most $\delta$ over $\Dcal\sim p^\star$, 
    \begin{align*}
        -\log(p^\star(\Dcal)/p(\Dcal))+
        \textup{\texttt{H}}^2(p^\star, p)\ge 2\log(1/\delta).
    \end{align*}
Thus, with probability at least $1-\delta$ over $\Dcal\sim p^\star$, 
    \begin{align*}
        -\log(p^\star(\Dcal)/p(\Dcal))+
        \textup{\texttt{H}}^2(p^\star, p)\le 2\log(1/\delta).
    \end{align*}
    Finally, taking union bounds on $p\in\Pcal$ yields that, 
    with probability at least $1-\delta$ on $\Dcal\sim p^\star$, for all $p\in\Pcal$, 
    \begin{align*}
        -\log(p^\star(\Dcal)/p(\Dcal))+
        \textup{\texttt{H}}^2(p^\star, p)\le 2\log(|\Pcal|/\delta).
    \end{align*}
    Since the above argument applies to any $p^\star\in\Pcal$, we have that for any $p^\star\in\Pcal$, 
    with probability at least $1-\delta$ over $\Dcal\sim p^\star$, for all $p\in\Pcal$, 
    \begin{align*}
        -\log(p^\star(\Dcal)/p(\Dcal))+
    \textup{\texttt{H}}^2(p^\star, p)\le 2\log(|\Pcal|/\delta).
    \end{align*}
    This completes the proof.
\end{proof}

\subsection{Proof of \Cref{theorem:up1}}

Fix the DMOF $(\Mcal, \Pi, \Ocal, \Lcal)$ and $M^\star\in\Mcal$.
We further denote the log-likelihood ratio of a model $M\in\Mcal$ w.r.t. the dataset $\Dcal$ as 
\begin{align*}
    \lli(\Dcal, M)\coloneqq
    \log(\mathbb{P}_{M^\star}(\Dcal))-\log(\mathbb{P}_{M}(\Dcal)).
\end{align*}
We have that 
\begin{align}
    \lli(\Dcal, M^\star)=0 
\end{align}
for any $\Dcal\in\Ocal$, which directly comes from the definition.
Since $\log(\mathbb{P}_{M^\star}(\Dcal^\star))$ is fixed given $M^\star$ and $\Dcal^\star$, EDD~(\Cref{eqn:ed2}) is equivalent to taking
\begin{align*}
    \hat{p}(\Dcal^\star)=\argmin_{p\in\Delta(\Pi)}\max_{M\in\Mcal} \Big[L(M, p)-\lambda\cdot\lli(\Dcal^\star, M)\Big].
\end{align*}
Thus, the loss can be written as 
\begin{align*}
    &\Lcal(M^\star, \hat{p}(\Dcal^\star))\\=&
    \Lcal(M^\star, \hat{p}(\Dcal^\star))-\lambda\cdot\lli(\Dcal^\star, M^\star)+\lambda\cdot\lli(\Dcal^\star, M^\star)\\=&
    \Lcal(M^\star, \hat{p}(\Dcal^\star))-\lambda\cdot\lli(\Dcal^\star, M^\star)\\\le&
    \max_{M\in\Mcal}\Big[\Lcal(M, \hat{p}(\Dcal^\star))-\lambda\cdot\lli(\Dcal^\star, M)\Big].
\end{align*}
By the definition of EDD and $\eoec$, we have that
\begin{align*}
    \Lcal(M^\star, \hat{p}(\Dcal^\star))\le&
    \max_{M\in\Mcal}\Big[\Lcal(M, \hat{p}(\Dcal^\star))-\lambda\cdot\lli(\Dcal^\star, M)\Big]\\=&
    \min_{p\in\Delta(\Pi)}\max_{M\in\Mcal}\Big[\Lcal(M, p)-\lambda\cdot\lli(\Dcal^\star, M)\Big]\\=&
    \min_{p\in\Delta(\Pi)}\max_{M\in\Mcal}\Big[\Lcal(M, p)+\lambda\log(\mathbb{P}_{M}(\Dcal^\star)/\mathbb{P}_{M^\star}(\Dcal^\star))\Big]\\=&
    \eoec_{\lambda}(M^\star, \Mcal, \Dcal^\star).
\end{align*}

This completes the proof.

\subsection{Proof of \Cref{theorem:up2}}

Fix $\lambda\ge 0$ and $M\in\Mcal$. %
By the definition of the $\eoec$ and \Cref{lemma:ll}, with probability at least $1-\delta$ over $\Dcal\sim M$, 
\begin{align*}
    \eoec_{\lambda}(M, \Mcal, \Dcal)=&\min_{p\in\Delta(\Pi)}\max_{M^\prime\in\Mcal} \big[\Lcal(M^\prime, p) -\lambda\cdot\log(\mathbb{P}_{M}(\Dcal)/ \mathbb{P}_{M^\prime}(\Dcal))\big]\Big]\\\le&
    \min_{p\in\Delta(\Pi)}\max_{M^\prime\in\Mcal} \big[\Lcal(M^\prime, p) -\lambda\cdot\HH(M, M^\prime)\big]+2\lambda\cdot\log(|\Mcal|/\delta)\\=&
    \oec_{\lambda}(\Mcal, \{M\})+2\lambda\cdot\log(|\Mcal|/\delta), 
\end{align*}
where the third equality follows from the definition of $\oec$. 
After this, \Cref{eqn:ub-m} is simply a consequence of \Cref{theorem:up1}.
This completes the proof.

\section{Missing Proofs in \Cref{section:fast}}
\label{proof:fast}

\subsection{Helper Lemmas}

\begin{lemma}[Subadditivity for the Squared Hellinger Distance; Trimmed Lemma A.13 in \citet{foster2023statistical}]
    \label{lemma:subadd-h}
    Let $\{(\Xcal_i, \Fcal_i)\}_{i\in[n]}$ be a sequence of measurable spaces, $\Xcal^{(i)}=\prod_{t=1}^i\Xcal_t$, and $\Fcal^{(i)}=\bigotimes_{t=1}^i\Fcal_i$.
    For each $i\in[n]$, let $\mathbb{P}^{(i)}(\cdot|\cdot)$ and $\mathbb{Q}^{(i)}(\cdot|\cdot)$ be probability kernels from $(\Xcal^{(i-1)}, \Fcal^{(i-1)})$ to $(\Xcal_i, \Fcal_i)$.
    Let $\mathbb{P}$ and $\mathbb{Q}$ be the laws of $\{X_i\}_{i\in[n]}$ under $X_i\sim\mathbb{P}^{(i)}(\cdot|\{X_t\}_{t\in[0, i-1]})$ and 
    $X_i\sim\mathbb{Q}^{(i)}(\cdot|\{X_t\}_{t\in[0, i-1]})$ respectively.
    Then it holds that
    \begin{align*}
        \HH(\mathbb{P}, \mathbb{Q})\le 10^2\log(n)\cdot\mathbb{E}_{\mathbb{P}}\Big[\sum_{i=1}^n \HH(\mathbb{P}(\cdot|\{X_t\}_{t\in[i-1]}), \mathbb{Q}(\cdot|\{X_t\}_{t\in[i-1]})\Big].
    \end{align*}
\end{lemma}

\begin{proof}
    See \citet{foster2023statistical}.
\end{proof}

\begin{lemma}[Union Bound of the Likelihood Under I.I.D.]
    \label{lemma:ll-iid}
    Let $\Pcal$ be a set of probabilities, and $\Dcal=\{X_i\}_{i\in[N]}$ be the i.i.d. dataset drawn from $p^{\otimes N}$ where $p\in\Pcal$. 
    For any $p\in\Pcal$, with probability at least $1-\delta$ over $\Dcal\sim p^{\otimes N}$, for all $p^\prime\in\Pcal$, 
    \begin{align*}
        \sum_{X_i\in\Dcal}\Big[-\log(p(X_i)/p^\prime(X_i))+
        \textup{\texttt{H}}^2(p, p^\prime)\Big]\le 2\log(|\Pcal|/\delta).
    \end{align*}
\end{lemma}

\begin{proof}
    We first fix $p, p^\prime\in\Pcal$, and take 
    \begin{align*}
        L(\Dcal)\coloneqq
        -\frac{1}{2}\log(p^{\otimes N}(\Dcal)/(p^\prime)^{\otimes N}(\Dcal))
        -\log\mathbb{E}_{\Dcal^\prime\sim p^{\otimes N}}\big[\sqrt{(p^\prime)^{\otimes N}(\Dcal^\prime)/p^{\otimes N}(\Dcal^\prime)}\big]
    \end{align*}
    We have that 
    \begin{align*}
        \mathbb{E}_{\Dcal\sim p^{\otimes N}}\Big[\exp\big(L(\Dcal)\big)\Big]=&
        \mathbb{E}_{\Dcal\sim p^{\otimes N}}\Big[\exp\big(-\frac{1}{2}\log(p^{\otimes N}(\Dcal)/(p^\prime)^{\otimes N}(\Dcal))-
                \log{\mathbb{E}_{\Dcal^\prime\sim p^{\otimes N}}\Big[\sqrt{(p^\prime)^{\otimes N}(\Dcal^\prime)/p^{\otimes N}(\Dcal^\prime)}\Big]}
        \big)\Big]\\=&
        \frac{\mathbb{E}_{\Dcal\sim p^{\otimes N}}\Big[\exp\big(-\frac{1}{2}\log(p^{\otimes N}(\Dcal)/(p^\prime)^{\otimes N}(\Dcal))\big)\Big]}
        {\mathbb{E}_{\Dcal\sim p^{\otimes N}}\big[\sqrt{(p^\prime)^{\otimes N}(\Dcal)/p^{\otimes N}(\Dcal)}\big]}\\=&1
    \end{align*}
    We can then apply the Chernoff bound, which yields that 
    \begin{align}
        \mathbb{P}_{\Dcal\sim p^{\otimes N}}\big(L(\Dcal)\ge\epsilon\big)\le&
        \frac{\mathbb{E}_{\Dcal\sim p^{\otimes N}}\exp\big(L(\Dcal)\big)}{\exp(\epsilon)}=\exp(-\epsilon).\label{eqn:che-eqn-fast-1}
    \end{align}
    Further, by the definition of the squared Hellinger distance, 
    \begin{align*}
        &\mathbb{P}_{\Dcal\sim p^{\otimes N}}\Big(
        \sum_{X_i\in\Dcal}\Big[-\log(p(X_i)/p^\prime(X_i))+
        \textup{\texttt{H}}^2(p, p^\prime)\Big]\ge\epsilon
        \Big)\\\le&
        \mathbb{P}_{\Dcal\sim p^{\otimes N}}\Big(
        \sum_{X_i\in\Dcal}\Big[-\log(p(X_i)/p^\prime(X_i))+
            2-2\mathbb{E}_{X\sim p}\big[\sqrt{p^\prime(X)/p(X)}\big]
        \Big]\ge\epsilon
    \Big)\\\le&
    \mathbb{P}_{\Dcal\sim p^{\otimes N}}\Big(
        \sum_{X_i\in\Dcal}\Big[-\log(p(X_i)/p^\prime(X_i))+
            2-2\exp\log\mathbb{E}_{X\sim p}\big[\sqrt{p^\prime(X)/p(X)}\big]
        \Big]\ge\epsilon\Big).
    \end{align*}
    Due to the fact that $\exp(x)\ge x+1$ for $x\in\mathbb{R}$, we further have 
    \begin{align*}
        &\mathbb{P}_{\Dcal\sim p^{\otimes N}}\Big(
        \sum_{X_i\in\Dcal}\Big[-\log(p(X_i)/p^\prime(X_i))+
        \textup{\texttt{H}}^2(p, p^\prime)\Big]\ge\epsilon
        \Big)\\\le&
        \mathbb{P}_{\Dcal\sim p^{\otimes N}}\Big(
        \sum_{X_i\in\Dcal}\Big[-\log(p(X_i)/p^\prime(X_i))
            -2\log\mathbb{E}_{X\sim p}\big[\sqrt{p^\prime(X)/p(X)}\big]
        \Big]\ge\epsilon
        \Big).
    \end{align*}
    On one hand, because of the i.i.d. assumption~(\Cref{assumption:sl-iid}), we obtain that 
    \begin{align*}
        \log\mathbb{E}_{\Dcal^\prime\sim p^{\otimes N}}\big[\sqrt{(p^\prime)^{\otimes N}(\Dcal^\prime)/p^{\otimes N}(\Dcal^\prime)}\big]=&
        \log\mathbb{E}_{\Dcal^\prime\sim p^{\otimes N}}\Bigg[\sqrt{\prod_{X_i\in\Dcal^\prime}p^\prime(X_i)/\prod_{X_i\in\Dcal^\prime}(p)(X_i)}\Bigg]\\=&
        \log\mathbb{E}_{\Dcal^\prime\sim p^{\otimes N}}\prod_{X_i\in\Dcal^\prime}\big[\sqrt{p^\prime(X_i)/(p)(X_i)}\big]\\=&
        N\log\mathbb{E}_{X\sim p}\big[\sqrt{p^\prime(X)/p(X)}\big]\\=&
        \sum_{X_i\in\Dcal^\prime}\log\mathbb{E}_{X\sim p}\big[\sqrt{p^\prime(X)/p(X)}\big].
    \end{align*}
    Thanks again to the i.i.d. assumption, we have that 
    \begin{align*}
        &\mathbb{P}_{\Dcal\sim p^{\otimes N}}\Big(
        \sum_{X_i\in\Dcal}\Big[-\log(p(X_i)/p^\prime(X_i))+
        \textup{\texttt{H}}^2(p, p^\prime)\Big]\ge\epsilon
        \Big)\\\le&
        \mathbb{P}_{\Dcal\sim p^{\otimes N}}\Big(
            \sum_{X_i\in\Dcal}\Big[-\log(p(X_i)/p^\prime(X_i))\Big]
            -2\log\mathbb{E}_{\Dcal^\prime\sim p^{\otimes N}}\big[\sqrt{(p^\prime)^{\otimes N}(\Dcal^\prime)/p^{\otimes N}(\Dcal^\prime)}\big]
        \ge\epsilon
    \Big)\\\le&
    \mathbb{P}_{\Dcal\sim p^{\otimes N}}\Big(
            -\log\Big(\prod_{X_i\in\Dcal}p(X_i)/\prod_{X_i\in\Dcal}(p^\prime)(X_i)\Big)
            -2\log\mathbb{E}_{\Dcal^\prime\sim p^{\otimes N}}\big[\sqrt{(p^\prime)^{\otimes N}(\Dcal^\prime)/p^{\otimes N}(\Dcal^\prime)}\big]
        \ge\epsilon
        \Big)\\\le&
        \mathbb{P}_{\Dcal\sim p^{\otimes N}}\Big(
            -\log(p^{\otimes N}(\Dcal)/(p^\prime)^{\otimes N}(\Dcal))
            -2\log\mathbb{E}_{\Dcal^\prime\sim p^{\otimes N}}\big[\sqrt{(p^\prime)^{\otimes N}(\Dcal^\prime)/p^{\otimes N}(\Dcal^\prime)}\big]
        \ge\epsilon
        \Big)\\\le&
        \mathbb{P}_{\Dcal\sim p^{\otimes N}}\Big(
            2L(\Dcal)\ge\epsilon
        \Big)\\\le&\exp(-\epsilon/2), 
    \end{align*}
    where the last inequality follows from \Cref{eqn:che-eqn-fast-1}.
    Taking $\delta=\exp(-\epsilon/2)$~(thus $\epsilon=2\log(1/\delta)$), we have that for  with probability at least $1-\delta$ over $\Dcal\sim p^{\otimes N}$, 
    \begin{align*}
        \sum_{X_i\in\Dcal}\Big[-\log(p(X_i)/p^\prime(X_i))+
        \textup{\texttt{H}}^2(p, p^\prime)\Big] \le 2\log(1/\delta).
    \end{align*}
    Finally, taking union bounds on $p^\prime\in\Pcal$ yields that, 
    with probability at least $1-\delta$ over $\Dcal\sim p^{\otimes N}$, for all $p^\prime\in\Pcal$, 
    \begin{align*}
        \sum_{X_i\in\Dcal}\Big[-\log(p(X_i)/p^\prime(X_i))+
        \textup{\texttt{H}}^2(p, p^\prime)\Big]\le 2\log(|\Pcal|/\delta).
    \end{align*}
    Note that the above arguments apply to all $p\in\Pcal$, which will yield the \Cref{lemma:ll-iid}.
    This completes the proof.
\end{proof}

\begin{lemma}[Union Bound of the Log-Likelihood Under I.I.D. with Conditioning]
    \label{lemma:ll-iid-c}
    Let $\Pcal$ be a set of probabilities on $\Xcal_1\times\Xcal_2$, 
    and $\Dcal=\{X_{1, n}\times X_{2, n}\}_{n\in[N]}$ be the i.i.d. dataset drawn from $p^{\otimes N}$ where $p\in\Pcal$. 
    For any $p\in\Pcal$, with probability at least $1-\delta$ over $\Dcal$, for all $p^\prime\in\Pcal$, 
    \begin{align*}
        \sum_{(X_1, X_2)\in\Dcal}\Big[-\log(p(X_2| X_1)/p^\prime(X_2| X_1))+
        \mathbb{E}_{X_1^\prime\sim p^\prime}\big[\textup{\texttt{H}}^2(p(\cdot| X_1^\prime), p^\prime(\cdot| X_1^\prime))\big]\Big]\le 2\log(|\Pcal|/\delta).
    \end{align*}
\end{lemma}

\begin{proof}

    Given the dataset $\Dcal=\{X_{1, n}\times X_{2, n}\}_{n\in[N]}$, we denote $\Dcal_1\coloneqq\{X_{1, n}\}_{n\in[N]}$ and $\Dcal_2\coloneqq\{X_{2, n}\}_{n\in[N]}$.
    We first fix $p, p^\prime\in\Pcal$, and define
    \begin{align*}
        L(\Dcal)\coloneqq
        -\frac{1}{2}\log(p^{\otimes N}(\Dcal_2|\Dcal_1)/(p^\prime)^{\otimes N}(\Dcal_2|\Dcal_1))
        -\log\mathbb{E}_{\Dcal^\prime\sim p^{\otimes N}}\big[\sqrt{(p^\prime)^{\otimes N}(\Dcal^\prime_2|\Dcal^\prime_1)/p^{\otimes N}(\Dcal^\prime_2|\Dcal^\prime_1)}\big].
    \end{align*}
    We then compute the moment generating function of the above function. By the definition, we have that 
    \begin{align*}
        &\mathbb{E}_{\Dcal\sim p^{\otimes N}}\Big[\exp\big(L(\Dcal)\big)\Big]\\=&
        \mathbb{E}_{\Dcal\sim p^{\otimes N}}\Big[\exp\big(
                -\frac{1}{2}\log(p^{\otimes N}(\Dcal_2|\Dcal_1)/(p^\prime)^{\otimes N}(\Dcal_2|\Dcal_1))-
                \log{\mathbb{E}_{\Dcal^\prime\sim p^{\otimes N}}\big[\sqrt{(p^\prime)^{\otimes N}(\Dcal_2^\prime|\Dcal_1^\prime)/p^{\otimes N}(\Dcal_2^\prime|\Dcal_1^\prime)}\big]}
        \big)\Big]\\=&
        \frac{\mathbb{E}_{\Dcal\sim p^{\otimes N}}\Big[\exp\big(
        -\frac{1}{2}\log(p^{\otimes N}(\Dcal_2|\Dcal_1)/(p^\prime)^{\otimes N}(\Dcal_2|\Dcal_1))\big)\Big]}{
        {\mathbb{E}_{\Dcal^\prime\sim p^{\otimes N}}\big[\sqrt{(p^\prime)^{\otimes N}(\Dcal_2^\prime|\Dcal_1^\prime)/p^{\otimes N}(\Dcal_2^\prime|\Dcal_1^\prime)}\big]}
\big)}\\=&
\frac{\mathbb{E}_{\Dcal\sim p^{\otimes N}}\Big[
\sqrt{(p^\prime)^{\otimes N}(\Dcal_2|\Dcal_1)/p^{\otimes N}(\Dcal_2|\Dcal_1)}\Big]}{
{\mathbb{E}_{\Dcal^\prime\sim p^{\otimes N}}\big[\sqrt{(p^\prime)^{\otimes N}(\Dcal_2^\prime|\Dcal_1^\prime)/p^{\otimes N}(\Dcal_2^\prime|\Dcal_1^\prime)}\big]}
\big)}\\=&1, 
    \end{align*}
    where the second equation follows from the fact that the latter term is a constant.
    We can then apply the Chernoff bound, which yields that 
    \begin{align}
        \mathbb{P}_{\Dcal\sim p^{\otimes N}}\big(L(\Dcal)\ge\epsilon\big)\le&
        \frac{\mathbb{E}_{\Dcal\sim p^{\otimes N}}\exp\big(L(\Dcal)\big)}{\exp(\epsilon)}=\exp(-\epsilon).\label{eqn:che-eqn-fast-2}
    \end{align}
    The following part of the proof is devoted to connecting $L$ and the equation in the lemma.
    By the definition of the squared Hellinger distance, 
    \begin{align}
        &\mathbb{P}_{\Dcal\sim p^{\otimes N}}\Bigg(
        \sum_{(X_1, X_2)\in\Dcal}\Big[-\log(p(X_2|X_1)/p^\prime(X_2|X_1))+
        \mathbb{E}_{X_1^\prime\sim p^\prime}\textup{\texttt{H}}^2(p(\cdot|X_1^\prime), p^\prime(\cdot|X_1^\prime))\Big]\ge\epsilon
        \Bigg)\\=&
        \mathbb{P}_{\Dcal\sim p^{\otimes N}}\Bigg(
        \sum_{(X_1, X_2)\in\Dcal}\Big[-\log(p(X_2|X_1)/p^\prime(X_2|X_1))+
            2-2\mathbb{E}_{(X_1^\prime, X_2^\prime)\sim p}\big[\sqrt{p^\prime(X_2^\prime|X_1^\prime)/p(X_2^\prime|X_1^\prime)}\big]
        \Big]\ge\epsilon
        \Bigg)\\=&
        \mathbb{P}_{\Dcal\sim p^{\otimes N}}\Bigg(
        \sum_{(X_1, X_2)\in\Dcal}\Big[-\log(p(X_2|X_1)/p^\prime(X_2|X_1))+
            2-2\exp\log\mathbb{E}_{(X_1^\prime, X_2^\prime)\sim p}\big[\sqrt{p^\prime(X_2^\prime|X_1^\prime)/p(X_2^\prime|X_1^\prime)}\big]
        \Big]\ge\epsilon
        \Bigg)\\\le&
        \mathbb{P}_{\Dcal\sim p^{\otimes N}}\Bigg(
        \sum_{(X_1, X_2)\in\Dcal}\Big[-\log(p(X_2|X_1)/p^\prime(X_2|X_1))
            -2\log\mathbb{E}_{(X_1^\prime, X_2^\prime)\sim p}\big[\sqrt{p^\prime(X_2^\prime|X_1^\prime)/p(X_2^\prime|X_1^\prime)}\big]
        \Big]\ge\epsilon
    \Bigg)\\=&
        \mathbb{P}_{\Dcal\sim p^{\otimes N}}\Bigg(
    \sum_{(X_1, X_2)\in\Dcal}\Big[-\log(p(X_2|X_1)/p^\prime(X_2|X_1))\Big]
            -2N\log\mathbb{E}_{(X_1^\prime, X_2^\prime)\sim p}\big[\sqrt{p^\prime(X_2^\prime|X_1^\prime)/p(X_2^\prime|X_1^\prime)}\big]
        \ge\epsilon
    \Bigg), \label{eqn:H2-log}
    \end{align}
    where the last inequality follows from the fact $\exp(x)\ge x+1$ for all $x\in\mathbb{R}$.
    On the other hand, as a result of the i.i.d. assumption, we can separate the multiplication from the expectation: 
    \begin{align*}
        &\log\mathbb{E}_{\Dcal^\prime\sim p^{\otimes N}}\big[\sqrt{(p^\prime)^{\otimes N}(\Dcal_2^\prime|\Dcal_1^\prime)/p^{\otimes N}(\Dcal^\prime_2|\Dcal_1^\prime)}\big]\\=&
        \log\mathbb{E}_{\Dcal^\prime\sim p^{\otimes N}}\Bigg[\sqrt{\prod_{(X_1^\prime, X_2^\prime)\in\Dcal^\prime}\Big(p^\prime(X_2^\prime| X_1^\prime)/p(X_2^\prime|X_1^\prime)}\Big)\Bigg]\\=&
        \log\mathbb{E}_{\Dcal^\prime\sim p^{\otimes N}}\prod_{(X_1^\prime, X_2^\prime)\in\Dcal^\prime}\Bigg[\sqrt{\Big(p^\prime(X_2^\prime| X_1^\prime)/p(X_2^\prime|X_1^\prime)}\Big)\Bigg]\\=&
        N\log\mathbb{E}_{(X_1^\prime, X_2^\prime)\sim p}\big[\sqrt{p^\prime(X_2^\prime|X_1^\prime)/p(X_2^\prime|X_1^\prime)}\big].
    \end{align*}
    As a conclusion of \Cref{eqn:H2-log} and the above equation, we have that
    \begin{align*}
        &\mathbb{P}_{\Dcal\sim p^{\otimes N}}\Big(
        \sum_{(X_1, X_2)\in\Dcal}\Big[-\log(p(X_2|X_1)/p^\prime(X_2|X_1))+
        \mathbb{E}_{X_1^\prime\sim p^\prime}\textup{\texttt{H}}^2(p(\cdot|X_1^\prime), p^\prime(\cdot|X_1^\prime))\Big]\ge\epsilon
        \Big)\\\le&
        \mathbb{P}_{\Dcal\sim p^{\otimes N}}\Bigg(
        \sum_{(X_1, X_2)\in\Dcal}-\log(p(X_2|X_1)/p^\prime(X_2|X_1))
            -2\log\mathbb{E}_{\Dcal^\prime\sim p^n}\big[\sqrt{(p^\prime)^{\otimes N}(\Dcal_2^\prime|\Dcal_1^\prime)/p^{\otimes N}(\Dcal^\prime_2|\Dcal_1^\prime)}\big]
        \ge\epsilon
    \Bigg).
    \end{align*}
    Again we can apply the i.i.d. assumption, we can merge the additions in the first term in to one:
    \begin{align*}
        &\mathbb{P}_{\Dcal\sim p^{\otimes N}}\Bigg(
        \sum_{(X_1, X_2)\in\Dcal}-\log(p(X_2|X_1)/p^\prime(X_2|X_1))
            -2\log\mathbb{E}_{\Dcal^\prime\sim p^n}\big[\sqrt{(p^\prime)^{\otimes N}(\Dcal_2^\prime|\Dcal_1^\prime)/p^{\otimes N}(\Dcal^\prime_2|\Dcal_1^\prime)}\big]
        \ge\epsilon
    \Bigg)\\\le&
        \mathbb{P}_{\Dcal\sim p^{\otimes N}}\Big(
            -\log(p^{\otimes N}(\Dcal_2|\Dcal_1)/(p^\prime)^{\otimes N}(\Dcal_2|\Dcal_1))
            -2\log\mathbb{E}_{\Dcal^\prime\sim p^n}\big[\sqrt{(p^\prime)^{\otimes N}(\Dcal_2^\prime|\Dcal_1^\prime)/p^{\otimes N}(\Dcal^\prime_2|\Dcal_1^\prime)}\big]
        \ge\epsilon
        \Big)\\\le&
        \mathbb{P}_{\Dcal\sim p^{\otimes N}}\Big(
    2L(\Dcal)\ge\epsilon\Big)\\\le&
        \exp(-\epsilon), 
    \end{align*}
    where the last inequality follows from \Cref{eqn:che-eqn-fast-2}.
    Taking $\delta=\exp(-\epsilon/2)$, we have that for all $p^\prime\in\Pcal$, with probability at least $1-\delta$ on $\Dcal\sim p^{\otimes N}$, 
    \begin{align*}
        \sum_{(X_1, X_2)\in\Dcal}\Big[-\log(p(X_2|X_1)/p^\prime(X_2|X_1))+
        \mathbb{E}_{X_1^\prime\sim p^\prime}\big[\textup{\texttt{H}}^2(p(\cdot| X_1^\prime), p^\prime(\cdot| X_1^\prime))\big]\Big]\le
        2\log(1/\delta).
    \end{align*}
    Finally, taking union bounds on $p^\prime\in\Pcal$ yields that, 
    for any $p\in\Pcal$, with probability at least $1-\delta$ over $\Dcal\sim p^{\otimes N}$, for all $p^\prime\in\Pcal$, 
    \begin{align*}
        \sum_{(X_1, X_2)\in\Dcal}\Big[-\log(p(X_2| X_1)/p^\prime(X_2| X_1))+
        \mathbb{E}_{X_1^\prime\sim p^\prime}\big[\textup{\texttt{H}}^2(p(\cdot| X_1^\prime), p^\prime(\cdot| X_1^\prime))\big]\Big]\le 2\log(|\Pcal|/\delta).
    \end{align*}
    This completes the proof.
\end{proof}

\begin{lemma}[Refined Simulation Lemma]
    \label{lemma:r-sim}
    Given a Markovian sequential problem $(\Pcal$, $\Mcal$, $\{\Scal_h\}_{h\in[H]}$, $\{\Acal_h\}_{h\in[H]}$, $\Pi$, $\ell, N)$ we have that 
    \begin{align*}
        \Lcal(M, p)\le&
        3\Lcal(M^\prime, p)\\&\quad+4B\cdot
       10^2\log(2H)\cdot\mathbb{E}_{\{(s_h, a_h)\}_{h\in[H]}\sim\Tcal(\cdot| M, p)}\Big[\sum_{i=1}^H \HH(\Pcal(\cdot|M, h, s_{i-1}, a_{i-1}), \Pcal(\cdot|M^\prime, h, s_{i-1}, a_{i-1})\Big], 
    \end{align*}
    for any $M, M^\prime\in\Mcal$ and $p\in\Delta(\Pi)$, where $\Tcal(\cdot|M, p)$ denotes the probability over $\{(s_h, a_h)\}_{h\in[H]}$ that is sampled from $\Tcal(\cdot|M, \pi)$ where $\pi\sim p$.
\end{lemma}

\begin{proof}
    We first prove the case that $p$ is a singleton $\pi$.
    By \Cref{lemma:subadd-h}, 
    \begin{align*}
        &\HH(\Tcal(\cdot| M, \pi), \Tcal(\cdot| M^\prime, \pi))\\\le&10^2\log(2H)\cdot\mathbb{E}_{\{(s_h, a_h)\}_{h\in[H]}\sim\Tcal(\cdot| M, \pi)}\Big[\sum_{i=1}^H \HH(\Pcal(\cdot|M, h, \{(s_t, a_t)\}_{t\in[i-1]}), \Pcal(\cdot|M^\prime, h, \{s_t, a_t\}_{t\in[i-1]})\Big]
      \\&\quad\quad\quad+10^2\log(2H)\cdot\mathbb{E}_{\{(s_h, a_h)\}_{h\in[H]}\sim\Tcal(\cdot| M, \pi)}\Big[\sum_{i=1}^H \HH(\pi(\cdot| \{(s_t, a_t)\}_{t\in[i-1]}, s_i), \pi(\cdot|\{s_t, a_t\}_{t\in[i-1]}, s_i)\Big]\\=&
        10^2\log(2H)\cdot\mathbb{E}_{\{(s_h, a_h)\}_{h\in[H]}\sim\Tcal(\cdot| M, \pi)}\Big[\sum_{i=1}^H \HH(\Pcal(\cdot|M, h, \{(s_t, a_t)\}_{t\in[i-1]}), \Pcal(\cdot|M^\prime, h, \{s_t, a_t\}_{t\in[i-1]})\Big].
    \end{align*}
    Due to the Markovian property of $\Pcal$, we have that 
    \begin{align*}
        &10^2\log(2H)\cdot\mathbb{E}_{\{(s_h, a_h)\}_{h\in[H]}\sim\Tcal(\cdot| M, \pi)}\Big[\sum_{i=1}^H \HH(\Pcal(\cdot|M, h, \{(s_t, a_t)\}_{t\in[i-1]}), \Pcal(\cdot|M^\prime, h, \{s_t, a_t\}_{t\in[i-1]})\Big]\\=&
       10^2\log(2H)\cdot\mathbb{E}_{\{(s_h, a_h)\}_{h\in[H]}\sim\Tcal(\cdot| M, \pi)}\Big[\sum_{i=1}^H \HH(\Pcal(\cdot|M, h, s_{i-1}, a_{i-1}), \Pcal(\cdot|M^\prime, h, s_{i-1}, a_{i-1})\Big].
    \end{align*}
    As a consequence, 
    \begin{align*}
        &\HH(\Tcal(\cdot| M, \pi), \Tcal(\cdot| M^\prime, \pi))\\\le&
       10^2\log(2H)\cdot\mathbb{E}_{\{(s_h, a_h)\}_{h\in[H]}\sim\Tcal(\cdot| M, \pi)}\Big[\sum_{i=1}^H \HH(\Pcal(\cdot|M, h, s_{i-1}, a_{i-1}), \Pcal(\cdot|M^\prime, h, s_{i-1}, a_{i-1})\Big].
    \end{align*}
    Thus, thanks to \Cref{lemma:hh-change}, 
    \begin{align*}
        \Lcal(M, \pi)=&\mathbb{E}_{\{s_h, a_h\}_{h\in[H]}\sim\Tcal(\cdot|M, \pi)}\ell(\pi, \{s_h, a_h\})\\\le&
        3\mathbb{E}_{\{s_h, a_h\}_{h\in[H]}\sim\Tcal(\cdot|M^\prime, \pi)}\ell(\pi, \{s_h, a_h\})+4B\cdot\HH(\Tcal(\cdot| M, \pi), \Tcal(\cdot| M^\prime, \pi))\\\le&
        3\mathbb{E}_{\{s_h, a_h\}_{h\in[H]}\sim\Tcal(\cdot|M^\prime, \pi)}\ell(\pi, \{s_h, a_h\})\\&\quad\quad+4B\cdot
        10^2\log(2H)\cdot\mathbb{E}_{\{(s_h, a_h)\}_{h\in[H]}\sim\Tcal(\cdot| M, \pi)}\Big[\sum_{i=1}^H \HH(\Pcal(\cdot|M, h, s_{i-1}, a_{i-1}), \Pcal(\cdot|M^\prime, h, s_{i-1}, a_{i-1})\Big]\\=&
        3\Lcal(M^\prime, \pi)\\&\quad+4B\cdot
       10^2\log(2H)\cdot\mathbb{E}_{\{(s_h, a_h)\}_{h\in[H]}\sim\Tcal(\cdot| M, \pi)}\Big[\sum_{i=1}^H \HH(\Pcal(\cdot|M, h, s_{i-1}, a_{i-1}), \Pcal(\cdot|M^\prime, h, s_{i-1}, a_{i-1})\Big], 
    \end{align*}
    which means 
    \begin{align*}
        \Lcal(M, \pi)\le&
        3\Lcal(M^\prime, \pi)\\&\quad+4B\cdot
       10^2\log(2H)\cdot\mathbb{E}_{\{(s_h, a_h)\}_{h\in[H]}\sim\Tcal(\cdot| M, \pi)}\Big[\sum_{i=1}^H \HH(\Pcal(\cdot|M, h, s_{i-1}, a_{i-1}), \Pcal(\cdot|M^\prime, h, s_{i-1}, a_{i-1})\Big].
    \end{align*}
    Taking the expectation $\mathbb{E}_{\pi\sim p}$ on both sides yields the desired result.
    This completes the proof.
\end{proof}

\subsection{Missing Proof of \Cref{theorem:fast-sq}}

For any $M\in\Mcal$, let $\mathbb{P}_{M}$ denotes the law of data generating. Denoting $\Dcal^\star=\{(s_{n, h}, a_{n, h}, s^\prime_{n, h})\}_{h\in[H], n\in[N]}$, 
by the definition of $\eoec$ as well as the i.i.d. assumption of the dataset~(\Cref{assumption:s-iid}), we have that
\begin{align*}
    &\eoec_{\lambda}(M^\star, \Mcal, \Dcal^\star)\\=&
    \min_{p\in\Delta(\Pi)}\max_{M^\prime\in\Mcal}
        \big[\Lcal(M^\prime, p)
    +\lambda\cdot\log(\mathbb{P}_{M^\prime}(\Dcal^\star)/\mathbb{P}_{M^\star}(\Dcal^\star))\big]\\=&
    \min_{p\in\Delta(\Pi)}\max_{M^\prime\in\Mcal}
        \big[\Lcal(M^\prime, p)
        +\lambda\cdot\sum_{n\in[N]}\log(\mathbb{P}_{M^\prime}(\{(s_{n, h}, a_{n, h}, s^\prime_{n, h})\}_{h\in[H]})/\mathbb{P}_{M^\star}(\{(s_{n, h}, a_{n, h}, s^\prime_{n, h})\}_{h\in[H]}))\big].
\end{align*}
We can further expand the above equation by the definition of conditional probability.
\begin{align*}
    &\eoec_{\lambda}(M^\star, \Mcal, \Dcal^\star)\\=&
    \min_{p\in\Delta(\Pi)}\max_{M^\prime\in\Mcal}
        \big[\Lcal(M^\prime, p)
+\lambda\cdot\sum_{n\in[N]}\log(\mathbb{P}_{M^\prime}(\{(s_{n, h}, a_{n, h}, s^\prime_{n, h})\}_{h\in[H]})/\mathbb{P}_{M^\star}(\{(s_{n, h}, a_{n, h}, s^\prime_{n, h})\}_{h\in[H]}))\big]\\=&
    \min_{p\in\Delta(\Pi)}\max_{M^\prime\in\Mcal}
        \Big[\Lcal(M^\prime, p)
    \\\quad\quad&+\lambda\cdot\sum_{n\in[N]}\sum_{h\in[H]}\log(\mathbb{P}_{M^\prime}(s_{n, h}|\{s_{n, i}, a_{n, i}, s^\prime_{n, i}\}_{i\in[h-1]})/\mathbb{P}_{M^\star}(s_{n, h}|\{s_{n, i}, a_{n, i}, s^\prime_{n, i}\}_{i\in[h-1]}))
    \\\quad\quad&+\lambda\cdot\sum_{n\in[N]}\sum_{h\in[H]}\log(\mathbb{P}_{M^\prime}(a_{n, h}|\{s_{n, i}, a_{n, i}, s^\prime_{n, i}\}_{i\in[h-1]}, s_{n, h})/\mathbb{P}_{M^\star}(a_{n, h}|\{s_{n, i}, a_{n, i}, s^\prime_{n, i}\}_{i\in[h-1]}, s_{n, h}))
\\\quad\quad&+\lambda\cdot\sum_{n\in[N]}\sum_{h\in[H]}\log(\mathbb{P}_{M^\prime}(s^\prime_{n, h}|\{s_{n, i}, a_{n, i}, s^\prime_{n, i}\}_{i\in[h-1]}, s_{n, h}, a_{n,h})/\mathbb{P}_{M^\star}(s^\prime_{n, h}|\{s_{n, i}, a_{n, i}, s^\prime_{n, i}\}_{i\in[h-1]}, s_{n, h}, a_{n, h}))\Big].
\end{align*}
We then exchange the summation above, and get 
\begin{align*}
    &\eoec_{\lambda}(M^\star, \Mcal, \Dcal^\star)\\=&
    \min_{p\in\Delta(\Pi)}\max_{M^\prime\in\Mcal}
        \Big[\Lcal(M^\prime, p)
    \\\quad\quad&+\underbrace{\lambda\cdot\sum_{h\in[H]}\sum_{n\in[N]}\log(\mathbb{P}_{M^\prime}(s_{n, h}|\{s_{n, i}, a_{n, i}, s^\prime_{n, i}\}_{i\in[h-1]})/\mathbb{P}_{M^\star}(s_{n, h}|\{s_{n, i}, a_{n, i}, s^\prime_{n, i}\}_{i\in[h-1]}))}_{(\textup{I})}
    \\\quad\quad&+\underbrace{\lambda\cdot\sum_{h\in[H]}\sum_{n\in[N]}\log(\mathbb{P}_{M^\prime}(a_{n, h}|\{s_{n, i}, a_{n, i}, s^\prime_{n, i}\}_{i\in[h-1]}, s_{n, h})/\mathbb{P}_{M^\star}(a_{n, h}|\{s_{n, i}, a_{n, i}, s^\prime_{n, i}\}_{i\in[h-1]}, s_{n, h}))}_{(\textup{II})}
    \\\quad\quad&+\underbrace{\lambda\cdot\sum_{h\in[H]}\sum_{n\in[N]}\log(\mathbb{P}_{M^\prime}(s^\prime_{n, h}|\{s_{n, i}, a_{n, i}, s^\prime_{n, i}\}_{i\in[h-1]}, s_{n, h}, a_{n,h})/\mathbb{P}_{M^\star}(s^\prime_{n, h}|\{s_{n, i}, a_{n, i}, s^\prime_{n, i}\}_{i\in[h-1]}, s_{n, h}, a_{n, h}))}_{(\textup{III})}.
\end{align*}

At this point, we can now apply the concentration inequality.
Thanks to \Cref{lemma:ll-iid}, the i.i.d. assumption~(\Cref{assumption:s-iid}), and the union bound, with probability at least $1-\delta\cdot H$, 
\begin{align}
    &(\textup{I})\\\le&
-\lambda\cdot\sum_{h\in[H]}\sum_{n\in[N]}\mathbb{E}_{\{(s_i, a_i, s_i^\prime)\}_{i\in[h]}\sim\Phi(M^\star)\}}\HH(\mathbb{P}_{M^\star}(\cdot|\{s_{i}, a_{i}, s^\prime_{i}\}_{i\in[h-1]}), \mathbb{P}_{M^\prime}(\cdot|\{s_{i}, a_{i}, s^\prime_{i}\}_{i\in[h-1]}))+\lambda\log(|\Mcal|/\delta)\\\le&
\lambda\log(|\Mcal|/\delta).
\label{eqn:fast-p1}
\end{align}

Moreover, again due to \Cref{lemma:ll-iid-c}, the i.i.d. assumption~(\Cref{assumption:s-iid}) and the union bound, with probability at least $1-\delta\cdot H$, 
\begin{align}
    (\textup{II})\le&
-\lambda\cdot\sum_{h\in[H]}\sum_{n\in[N]}\mathbb{E}_{\{(s_i, a_i, s_i^\prime)\}_{i\in[h]}\sim\Phi(M^\star)\}}\Big[\\&\quad\quad\quad
\HH(\mathbb{P}_{M^\star}(\cdot|\{s_{i}, a_{i}, s^\prime_{i}\}_{i\in[h-1]}, s_{h}), \mathbb{P}_{M^\prime}(\cdot|\{s_{i}, a_{i}, s^\prime_{i}\}_{i\in[h-1]}, s_{h}))\Big]+\lambda\log(|\Mcal|/\delta)\\\le&\lambda\log(|\Mcal|/\delta).
\label{eqn:fast-p2}
\end{align}

For the same reason, we have that with probability at least $1-\delta \cdot H$, 
\begin{align}
    (\textup{III})\le&
-\lambda\cdot\sum_{h\in[H]}\sum_{n\in[N]}\mathbb{E}_{\{(s_i, a_i, s_i^\prime)\}_{i\in[h]}\sim\Phi(M^\star)\}}\Big[\\&\quad\quad\quad\HH(\mathbb{P}_{M^\star}(\cdot|\{s_{i}, a_{i}, s^\prime_{i}\}_{i\in[h-1]}, s_{h}, a_{h}), \mathbb{P}_{M^\prime}(\cdot|\{s_{i}, a_{i}, s^\prime_{i}\}_{i\in[h-1]}, s_{h}, a_{h}))\Big]\Big]+\lambda\log(|\Mcal|/\delta).
\label{eqn:fast-p3}
\end{align}

Combining \Cref{eqn:fast-p1,eqn:fast-p2,eqn:fast-p3} as well as the union bound yields that, with probability at least $1-\delta\cdot 3H$, 

\begin{align*}
    &\eoec_{\lambda}(M^\star, \Mcal, \Dcal^\star)\\\le&
    \min_{p\in\Delta(\Pi)}\max_{M^\prime\in\Mcal}
\Big[\Lcal(M^\prime, p)+3\lambda\log(|\Mcal|/\delta)-\lambda\cdot\sum_{h\in[H]}\sum_{n\in[N]}\mathbb{E}_{\{(s_i, a_i, s_i^\prime)\}_{i\in[h]}\sim\Phi(M^\star)\}}\Big[\\&\quad\quad\quad\HH(\mathbb{P}_{M^\star}(\cdot|\{s_{i}, a_{i}, s^\prime_{i}\}_{i\in[h-1]}, s_{h}, a_{h}), \mathbb{P}_{M^\prime}(\cdot|\{s_{i}, a_{i}, s^\prime_{i}\}_{i\in[h-1]}, s_{h}, a_{h}))\Big]\Big]\\=&
    \min_{p\in\Delta(\Pi)}\max_{M^\prime\in\Mcal}
\Big[\Lcal(M^\prime, p)+3\lambda\log(|\Mcal|/\delta)-\lambda N\cdot\sum_{h\in[H]}\mathbb{E}_{\{(s_i, a_i, s_i^\prime)\}_{h\in[H]}\sim\Phi(M^\star)\}}\Big[\\&\quad\quad\quad\HH(\mathbb{P}_{M^\star}(\cdot|\{s_{i}, a_{i}, s^\prime_{i}\}_{i\in[h-1]}, s_{h}, a_{h}), \mathbb{P}_{M^\prime}(\cdot|\{s_{i}, a_{i}, s^\prime_{i}\}_{i\in[h-1]}, s_{h}, a_{h}))\Big]\Big]\\=&
    \min_{p\in\Delta(\Pi)}\max_{M^\prime\in\Mcal}
\Big[\Lcal(M^\prime, p)+3\lambda\log(|\Mcal|/\delta)-\lambda N\cdot\sum_{h\in[H]}\mathbb{E}_{(s_h, a_h)\sim d^\Dcal_h(\cdot|M^\star)\}}\Big[\\&\quad\quad\quad\HH(\Pcal(\cdot|M^\star, h, s_{h}, a_{h}), \Pcal(\cdot|M^\prime, h, s_{h}, a_{h}))\Big]\Big].
\end{align*}
The subsequent arguments in this proof will be based on this high probability event.
Further due to the concentration assumption~(\Cref{assumption:coverage}), we have that 
\begin{align*}
    &\eoec_{\lambda}(M^\star, \Mcal, \Dcal^\star)\\\le&
    \min_{p\in\Delta(\Pi)}\max_{M^\prime\in\Mcal}
\Big[\Lcal(M^\prime, p)+3\lambda\log(|\Mcal|/\delta)-\frac{\lambda N}{C_{\bar{\pi}}}\cdot\sum_{h\in[H]}\mathbb{E}_{\{(s_h, a_h)\}_{h\in[H]}\sim \Tcal(\cdot|M^\star, \bar{\pi}, h)\}}\Big[\\&\quad\quad\quad\HH(\Pcal(\cdot|M^\star, h, s_{h}, a_{h}), \Pcal(\cdot|M^\prime, h, s_{h}, a_{h}))\Big]\Big]\\=&
    \min_{p\in\Delta(\Pi)}\max_{M^\prime\in\Mcal}
\Big[\Lcal(M^\prime, p)+3\lambda\log(|\Mcal|/\delta)-\frac{\lambda N}{C_{\bar{\pi}}}\cdot\sum_{h\in[H]}\mathbb{E}_{\{(s_h, a_h)\}_{h\in[H]}\sim \Tcal(\cdot|M^\star, \bar{\pi}, h)\}}\Big[\\&\quad\quad\quad\HH(\Pcal(\cdot|M^\prime, h, s_{h}, a_{h}), \Pcal(\cdot|M^\star, h, s_{h}, a_{h}))\Big]\Big].
\end{align*}

Setting $\lambda$ to $\frac{400BC_{\bar{\pi}}\log(2H)}{N}$, we can then apply the refined simulation lemma~(\Cref{lemma:r-sim}), 
we can conclude that 
\begin{align*}
    &\eoec_{\frac{400BC_{\bar{\pi}}\log(2H)}{N}}(M^\star, \Mcal, \Dcal^\star)\\\le&
    \min_{p\in\Delta(\Pi)}\max_{M^\prime\in\Mcal}
\Big[\Lcal(M^\prime, p)+\frac{1200BC_{\bar{\pi}}\log(2H)}{N}\log(|\Mcal|/\delta)\\\quad\quad\quad&-4B\cdot 10^2\log(2H)\cdot\sum_{h\in[H]}\mathbb{E}_{\{(s_h, a_h)\}_{h\in[H]}\sim \Tcal(\cdot|M^\star, \bar{\pi}, h)\}}\Big[\HH(\Pcal(\cdot|M^\prime, h, s_{h}, a_{h}), \Pcal(\cdot|M^\star, h, s_{h}, a_{h}))\Big]\Big]\\\le&
    \max_{M^\prime\in\Mcal}
\Big[\Lcal(M^\prime, \bar{\pi})+\frac{1200BC_{\bar{\pi}}\log(2H)}{N}\log(|\Mcal|/\delta)\\&\quad\quad\quad-4B\cdot 10^2\log(2H)\cdot\sum_{h\in[H]}\mathbb{E}_{\{(s_h, a_h)\}_{h\in[H]}\sim \Tcal(\cdot|M^\star, \bar{\pi}, h)\}}\Big[\HH(\Pcal(\cdot|M^\prime, h, s_{h}, a_{h}), \Pcal(\cdot|M^\star, h, s_{h}, a_{h}))\Big]\Big]\\\le&
3\Lcal(M^\star, \bar{\pi})+\frac{1200BC_{\bar{\pi}}\log(2H)}{N}\cdot\log(|\Mcal|/\delta).
\end{align*}
Setting the $\delta$ above to $\delta/(3H)$ can yield the first part of \Cref{theorem:fast-sq}.
On the other hand, \Cref{eqn:fast-sq2} is a consequence of the above equation and \Cref{theorem:up1}.
This completes the proof.

\section{Missing Proofs in \Cref{section:oec}}
\label{proof:oec}

\subsection{Proof for \Cref{theorem:lbexp}}
\label{subsection:proof-lbexp}

This subsection provides a proof for \Cref{theorem:lbexp}.
The proof is simple yet instructive, which mainly depends on two techniques:

\begin{itemize}
    \item the refined change of measure property of the squared Hellinger distance, i.e., the difference of the expected value of a function on a random variable
        can be bounded by the squared Hellinger distance~(\Cref{lemma:hh-change}); we summarize this property in \Cref{section:extension}.
    \item the trick from \citet{foster2023statistical}---if the distribution of the input of the algorithm is prefixed, we can 
        replace the algorithm's output with a term from $\Delta(\Pi)$.
\end{itemize}

\begin{proof}[Proof of \Cref{theorem:lbexp}]

    Fix $A\subseteq\Delta(\Ocal$).
    For any algorithm $\mathfrak{A}$, we can rewrite its worst case performance as 
    \begin{align}
        &\sup\limits_{M\in\Mcal} \mathbb{E}_{\Dcal\sim M}\big[\Lcal(M, \mathfrak{A}(\Dcal))\big]\\=&
        \label{eqn:d-trans}\sup\limits_{\rho\in A} \sup\limits_{M\in\Mcal} \mathbb{E}_{\Dcal\sim M}\big[\Lcal(M, \mathfrak{A}(\Dcal))\big]\\\ge&
        \frac{1}{3}\sup\limits_{\rho\in A} \sup\limits_{M\in\Mcal}\Big[\mathbb{E}_{\Dcal\sim \rho}\big[\Lcal(M, \mathfrak{A}(\Dcal))\big]-4B\cdot\HH(\rho, M)\Big], 
    \end{align}
    where the last inequality follows from \Cref{lemma:hh-change}, for which we take the whole $\Lcal(M, \mathfrak{A}(\cdot))$ as a function.
    We can then apply the trick from \citet{foster2023statistical}: \emph{as the algorithm only takes the dataset generated by $\rho$ as input, the distribution of the output policy would be fixed given $\rho$ and $\mathfrak{A}$.}
    More precisely, given $\Dcal\sim\rho$, let $\hat{\pi}$ denotes the output policy, we will have that 
    \begin{align*}
        \hat{\pi}\sim\mathfrak{A}(\Dcal)
    \end{align*}
    regardless of the choice of $M\in\Mcal$.
    This means that we can insert a $\min$ on the distribution of policies after $\sup_{\rho\in A}$.
    Denoting the distribution of $\hat{\pi}$ as $\hat{p}(\rho)$, we have that 
    \begin{align}
        &\sup\limits_{M\in\Mcal} \mathbb{E}_{\Dcal\sim M}\big[\Lcal(M, \mathfrak{A}(\Dcal))\big]\\\ge&
        \frac{1}{3}\sup\limits_{\rho\in A}\sup\limits_{M\in\Mcal}\Big[\Lcal(M, \hat{p}(\rho))-4B\cdot\HH(\rho, M)\Big]\\=&
        \label{eqn:a2p}\frac{1}{3}\sup\limits_{\rho\in A}\inf\limits_{p\in\Delta(\Pi)}\sup\limits_{M\in\Mcal}\Big[\Lcal(M, p)-4B\cdot\HH(\rho, M)\Big]\\=&
        \frac{1}{3}\oec_{4B}(\Mcal, A).
    \end{align}
    This completes the proof.
\end{proof}

\section{Obstacle of Applying Bracketing in Model-based Algorithms}
\label{section:bracketing}

\textcolor{black}{
In \citet{Uehara2021PessimisticMO}, the authors propose using the bracketing number to discretize the infinite function class, which allows us to apply results in function approximation to special cases like the tabular setting. One essential flaw of using the bracketing number in model-based methods is that it introduces an improper function class, i.e., a class that contains elements that are not probabilities~(e.g., $\tilde{\mathcal{M}}$ in Lemma 6 of \citet{Uehara2021PessimisticMO}). 
This flaw becomes problematic in their proof. Specifically, in their proof, they refer to \citet{agarwal2020flambe}, where the authors utilize the squared-Hellinger distance and use two forms of its definition:
\begin{align*}
    \HH(q\Vert p) = \int(\sqrt{p(z)} - \sqrt{q(z)})^2 dz \quad \textup{and} \quad \HH(q\Vert p) = 2 - \mathbb{E}_{z \sim q}\Big[ 1 - \sqrt{p(z) q(z)} \Big]. 
\end{align*} 
However, since one of the functions is essentially not a probability (or density), the two definitions contradict each other at a fundamental level. As a result, traditional discretization methods like bracketing and covering cannot be straightforwardly applied in model-based algorithms as they can in model-free ones. We may need to introduce new distance measures or assumptions, which could hopefully explain the mismatch in the order $N$ with special cases like tabular settings and provide new insights into the problem of sample complexity in offline RL.}

\section{$\oec$ with Other $f$-divergences}
\label{section:extension}

This section shows that one can replace the squared Hellinger distance in $\oec$ with another divergence that satisfies a refined change of measure property and still attain a similar lower bound result.
We first state the refined change of measure property in the following.

\begin{definition}[Refined Change of Measure Property]
\label{definition:ed}
We say that an $f$-divergence $D$ satisfies the refined change of measure property
with coefficients $\gamma_{D, 1}$ and $\gamma_{D, 2}$ if 
\begin{align*}
    \mathbb{E}_{X\sim \mathbb{P}}[g(X)]\ge \gamma_{D, 1}\Big[\mathbb{E}_{X\sim\mathbb{Q}}[g(X)]-\gamma_{D, 2}\cdot B\cdot D(\mathbb{P}, \mathbb{Q})\Big].
\end{align*}
for any function $g$ with range $[0, B]$, 
\end{definition}

Lots of the classic $f$-divergences satisfy the above property:
\begin{itemize}
    \item the total variation distance with $\gamma_{\TV, 1}=1$ and $\gamma_{\TV, 2}=2$~(from \Cref{lemma:tv-change}).
    \item the squared Hellinger distance with $\gamma_{\HH, 1}=\frac{1}{3}$ and $\gamma_{\HH, 2}=4$~(from \Cref{lemma:hh-change}).
    \item the KL divergence with $\gamma_{\KL, 1}=\frac{1}{3}$ and $\gamma_{\KL, 2}=4$~(from \Cref{lemma:hh-change} and the fact that $\KL(\mathbb{P}, \mathbb{Q})\ge\HH(\mathbb{P}, \mathbb{Q})$).
\end{itemize}

With an $f$-divergence $D$ satisfying the refined change of measure property, we can further define the generalized offline estimation coefficient as follows.

\begin{definition}
    \label{definition:extended-oec}
    Given a DMOF $(\Mcal, \Pi, \Ocal, \Lcal)$ and $f$-divergence $D$ satisfying the refined change of measure property~(\Cref{definition:ed}), \emph{its} $\oec$ w.r.t. a set of probability $A\subseteq\Delta(\Ocal)$ and a parameter $\lambda\ge 0$ is defined as 
    \begin{align}
        \label{eqn:extended-oec}
        \oec_{\lambda, D}(\Mcal, A) \coloneqq
        \sup\limits_{\rho\in A}\min_{p\in\Delta(\Pi)}\sup\limits_{M\in\Mcal}\Big[\Lcal(M, p)- B\cdot D(\rho, M)\Big].
    \end{align}
\end{definition}

\begin{theorem}
\label{theorem:extend-lbexp}
Given a DMOF $(\Mcal, \Pi, \Ocal, \Lcal)$, and an $f$-divergence $D$ that satisfies the refined change of measure property~(\Cref{definition:ed}), 
for any algorithm $\mathfrak{A}$, there exists an $M\in\Mcal$ such that the expected loss of the algorithm under model $M$ is lower bounded as follow: 
\begin{align}
    \label{eqn:lb-general-f}
    \mathbb{E}_{\Dcal\sim M}\Big[\Lcal(M, \mathfrak{A}(\Dcal))\Big]\ge\gamma_{D, 1}\cdot\oec_{\gamma_{D, 2}\cdot B, D}(\Mcal, A), 
\end{align}
where $A$ can be chosen as any subset of $\Delta(\Ocal)$.
\end{theorem}

\begin{proof}
    Fix $A\subseteq\Delta(\Ocal)$.
    For any algorithm $\mathfrak{A}$, we can rewrite its worst case performance as 
    \begin{align}
        &\sup\limits_{M\in\Mcal} \mathbb{E}_{\Dcal\sim M}\big[\Lcal(M, \mathfrak{A}(\Dcal))\big]\\=&
        \label{eqn:d-trans}\sup\limits_{\rho\in A} \sup\limits_{M\in\Mcal} \mathbb{E}_{\Dcal\sim M}\big[\Lcal(M, \mathfrak{A}(\Dcal))\big]\\\ge&
        \gamma_{D, 1}\sup\limits_{\rho\in A} \sup\limits_{M\in\Mcal}\Big[\mathbb{E}_{\Dcal\sim\rho}\Lcal(M, \mathfrak{A}(\Dcal))-\gamma_{D, 2}\cdot B\cdot D(\rho, M)\Big], 
    \end{align}
    where the last inequality follows from \Cref{definition:ed}, for which we take the whole $\Lcal(M, \mathfrak{A}(\cdot))$ as a function.
    We can apply the same trick used in \Cref{subsection:proof-lbexp}, and get 
    \begin{align}
        &\sup\limits_{M\in\Mcal} \mathbb{E}_{\Dcal\sim M}\big[\Lcal(M, \mathfrak{A}(\Dcal))\big]\\\ge&
        \gamma_{D, 1}\sup\limits_{\rho\in A} \sup\limits_{M\in\Mcal}\Big[\mathbb{E}_{\Dcal\sim\rho}\Lcal(M, \mathfrak{A}(\Dcal))-\gamma_{D, 2}\cdot B\cdot D(\rho, M)\Big]\\\ge&
        \gamma_{D, 1}\sup\limits_{\rho\in A}\inf_{p\in\Delta(\Pi)}\sup\limits_{M\in\Mcal}\Big[\Lcal(M, p)-\gamma_{D, 2}\cdot B\cdot D(\rho, M)\Big]\\\ge&
        \gamma_{D, 1}\oec_{\gamma_{D, 2}\cdot B, D}(\Mcal, A).
    \end{align}
    This completes the proof.
\end{proof}

\section{Extension Case: Supervised Learning and Sample-based Full Correlation}
\label{section:fast-sl}

In this section, we shows that in the case of considering \emph{supervised learning}~\citep{Vapnik1998,books/daglib/0033642} in the DMOF framework, EDD~(\Cref{eqn:ed2}) achieves a fast convergence rate compared with the optimal loss.

In supervised learning, there is an input space $\Xcal_{\SL}$, an output $\Ycal_{\SL}$, a dataset $\Dcal=\{(x_{\SL, i}, y_{\SL, i}\}_{i\in[N]}$\footnote{$[N]$ denotes the set of integers $\{1, 2, 3, \dots, N\}$.} that contains i.i.d. random variables from the distribution $d\in\Delta(\Xcal_{\SL}\times\Ycal_{\SL})$. While the distribution $d$ is \emph{unknown}, we are given a hypothesis class $\Hcal\subseteq(\Xcal_{\SL}\to\Ycal_{\SL})$ together with a dataset $\Dcal$ and wish to find a $\hat{h}\in\Hcal$ that minimizes
\begin{align}
    \mathbb{E}_{\Dcal}\sum_{(x_{\SL, i}, y_{\SL, i})\in\Dcal}L(\hat{h}(x_{\SL, i}), y_{\SL, i})=\mathbb{E}_{(X_{\SL}, Y_{\SL})\sim d} L(\hat{h}(X_{\SL}), Y_{\SL}), 
\end{align}
where $L\in(\Ycal_{\SL}\times\Ycal_{\SL}\to[0, B])$ is the loss function.

Supervised learning is a special instance of the DMOF with full correlation. We explain supervised learning with the DMOF in the following definition.

\begin{definition}[Supervised Learning]
\label{definition:sl}
Let $\Xcal$ be a feature space, $\ell:\Xcal\times\Pi\to[0, B]$ be a point-wise loss function, and $N$ be the size of the dataset. We can overload the DMOF~$(\Mcal, \Pi, \Ocal, \Lcal)$ by letting $\Ocal\coloneqq\Xcal^N$ be the set of ensembles of samples, and $\Mcal\subseteq\Delta(\Xcal^N)$ be a set of data distributions. 
We define the loss function as 
\begin{align*}
    \Lcal(M, \pi)\coloneqq \mathbb{E}_{\Dcal\sim M}\Big[\frac{1}{N}\sum_{x\in\Dcal} \ell(x, \pi)\Big].
\end{align*}
We refer to this framework as the DMOF \emph{over the supervised learning problem}~($\Xcal$, $\ell$, $N$).
\end{definition}

\begin{remark}
    Here we explain the correspondence of the above definition and supervised learning:
    the space $\Xcal=\Xcal_{\SL}\times\Ycal_{\SL}$ consists of two parts~(the input space and the output space), %
    $\Pi=\Hcal$ is the hypothesis class, and $\ell(x, \pi)=L(\pi(x_{\SL}), y_{\SL})$ is the sample-based loss function.
\end{remark}

Like supervised learning, we can make the i.i.d. assumption for DMOFs over supervised learning problems. 

\begin{assumption}[I.I.D. Assumption]
    \label{assumption:sl-iid}
    Given a DMOF $(\Mcal, \Pi, \Ocal, \Lcal)$ over a supervised learning problem~($\Xcal$, $\ell$, $N$), %
    the i.i.d. assumption assumes that any $M\in\Mcal$ is an $N$-letter product distribution of $\Phi(M)$~(i.e., $M=\Phi(M)^{\otimes N}$) where $\Phi(M)\in\Delta(\Xcal)$ can be any distribution over $\Xcal$. Namely, for any $M\in\Mcal$, its dataset $\Dcal=\{x_{n}\}_{n\in[N]}$ is drawn in an i.i.d. manner from $\Phi(M)$.
\end{assumption}

Combining the correlation introduced by supervised learning and the i.i.d. assumption, we can obtain our first fast convergence result, as presented in the following theorem.

\begin{theorem}
    \label{theorem:fast-sl}
    Given a DMOF $(\Mcal, \Pi, \Ocal, \Lcal)$ over a supervised learning problem ($\Xcal$, $\ell$, $N$) that meets \Cref{assumption:sl-iid}, for any $M\in\Mcal$, with probability at least $1-\delta$ over $\Dcal\sim M$, 
    we have
    \begin{align}
        \eoec_{4B/N}(M, \Mcal, \Dcal)\le&
    3\min_{\pi\in\Pi} \big[\Lcal(M, \pi)\big]+\frac{8B}{N}\cdot\log(|\Mcal|/\delta).
    \label{eqn:fast-sl1}
    \end{align}
    As a consequence, by choosing $\lambda$ as $4B/N$, EDD has its loss bounded, with probability at least $1-\delta$ over $\Dcal^\star\sim M^\star$, as 
    \begin{align}
        \Lcal(M^\star, \hat{p})\le&
        3\min_{\pi\in\Pi} \big[\Lcal(M^\star, \pi)\big]+\frac{8B}{N}\cdot\log(|\Mcal|/\delta).\label{eqn:fast-sl}
    \end{align}
\end{theorem}

There is a blowing up of the loss in the right hand side of \Cref{eqn:fast-sl}.
However, as long as there is a $\pi\in\Pi$ that gives a small value of $\Lcal(M^\star, \pi)$, this bound would be valuable. 
Moreover, this blowing up can be easily mitigated by subtracting the optimal loss from $\Lcal(M, \pi)$ in the algorithm.

\begin{proposition}
    \label{prop:1}
    Under the same context as \Cref{theorem:fast-sl}, by overloading the function $\Lcal(M, \pi)$ by $\Lcal(M, \pi)-\min_{\pi^\prime\in\Pi}\Lcal(M, \pi^\prime)$ in EDD, choosing $\lambda$ as $4B/N$ ensures with probability at least $1-\delta$ over $\Dcal^\star\sim M^\star$ that, EDD has its loss bounded as 
    \begin{align}
        \Lcal(M^\star, \hat{p})- \min_{\pi\in\Pi} \big[\Lcal(M^\star, \pi)\big]\le\frac{8B}{N}\cdot\log(|\Mcal|/\delta).
    \end{align}
\end{proposition}

The only crucial point to prove \Cref{prop:1} is to show $\Lcal(M, \pi)-\min_{\pi^\prime\in\Pi}\Lcal(M, \pi^\prime)\in[0, B]$, which is straightforward from the definition. With this, we can replace the $\Lcal$ in \Cref{theorem:fast-sl} and get \Cref{prop:1}.

\paragraph{Fast Convergence Rate in Supervised Learning} 
The fast convergence rate of supervised learning has been built in various ways~\citep{Vapnik1998,NIPS2013_a97da629}, and thus the result is not surprising. 
Nevertheless, as the focus of this paper is the partially correlated offline decision making, we think that \Cref{theorem:fast-sl} is inspiring and shows the power of the simple algorithm EDD.

\subsection{Proof of \Cref{theorem:fast-sl}}

    We first upper bound $\eoec$ with the use of concentration inequalities. Let $\mathbb{P}_M$ be the law of the data generating process under model $M$. By the definition $\eoec$ and the i.i.d. assumption~(\Cref{assumption:sl-iid}), 
\begin{align*}
    &\eoec_{\lambda}(M, \Mcal, \Dcal)\\=&
    \min_{p\in\Delta(\Pi)}\max_{M^\prime\in\Mcal}
        \big[\Lcal(M^\prime, p)
    +\lambda\cdot\log(\mathbb{P}_{M^\prime}(\Dcal)/\mathbb{P}_{M}(\Dcal))\big]\\\le&
    \min_{p\in\Delta(\Pi)}\max_{M^\prime\in\Mcal}
        \big[\Lcal(M^\prime, p)
+\lambda\cdot\sum_{n\in[N]}\log(\mathbb{P}_{M^\prime}(X_n)/\mathbb{P}_{M}(X_n))\big]\\=&
    \min_{p\in\Delta(\Pi)}\max_{M^\prime\in\Mcal}
        \big[\Lcal(M^\prime, p)
+\lambda\cdot\sum_{n\in[N]}\log(\Phi(M^\prime)(X_n)/\Phi(M)(X_n))\big].
\end{align*}

By \Cref{lemma:ll-iid} as well as again the i.i.d. assumption~(\Cref{assumption:sl-iid}), with probability at least $1-\delta$,

\begin{align*}
    &\eoec_{\lambda}(M, \Mcal, \Dcal)\\\le&
    \min_{p\in\Delta(\Pi)}\max_{M^\prime\in\Mcal}
        \Big[\Lcal(M^\prime, p)
    -\lambda\cdot N\cdot\HH(\Phi(M^\prime), \Phi(M))\Big]+2\lambda\cdot\log(|\Mcal|/\delta).
\end{align*}

We define $\ell(x, p)\coloneqq\mathbb{E}_{\pi\sim p}\ell(x, \pi)$. 
Due to the definition of $\Lcal$ and the i.i.d. assumption~(\Cref{assumption:sl-iid}), 
\begin{align*}
    \Lcal(M^\prime, p)=\mathbb{E}_{\Dcal^\prime\sim M^\prime}\Big[\frac{1}{N} \sum_{x\sim\Dcal^\prime}\ell(x, p)\Big] =
    \mathbb{E}_{x\sim \Phi(M^\prime)}\Big[\ell(x, p)\Big] .
\end{align*}

Setting $\lambda$ to $4B/N$ yields that 
\begin{align*}
    \eoec_{\lambda}(M, \Mcal, \Dcal)\le&
    \min_{p\in\Delta(\Pi)}\max_{M^\prime\in\Mcal}
    \Big[\mathbb{E}_{x\sim \Phi(M^\prime)}\Big[\ell(x, p)\Big] 
-4B\cdot\HH(\mathbb{P}_{M^\prime}, \mathbb{P}_{M})\Big]+\frac{8B}{N}\cdot\log(|\Mcal|/\delta)\\\le&
    \min_{p\in\Delta(\Pi)}\max_{M^\prime\in\Mcal}
    \Big[3\mathbb{E}_{x\sim \Phi(M)}\big[\ell(x, p)\big] \Big]+\frac{8B}{N}\cdot\log(|\Mcal|/\delta)\\\le&
    \min_{p\in\Delta(\Pi)}\max_{M^\prime\in\Mcal}
\big[3\Lcal(M, p)\big]+\frac{8B}{N}\cdot\log(|\Mcal|/\delta)\\=&
    \min_{p\in\Delta(\Pi)} \big[3\Lcal(M, p)\big]+\frac{8B}{N}\cdot\log(|\Mcal|/\delta)\\\le&
    \min_{\pi\in\Pi} \big[3\Lcal(M, \pi)\big]+\frac{8B}{N}\cdot\log(|\Mcal|/\delta).
\end{align*}
On other hand, the second part of \Cref{theorem:fast-sl} is a consequence of the above equation and \Cref{theorem:up1}.
This completes the proof.

\end{document}